\documentclass[runningheads]{llncs}

\usepackage{eccv}

\usepackage{eccvabbrv}

\usepackage[accsupp]{axessibility}  

\usepackage{url}
\usepackage{booktabs}       
\usepackage[utf8]{inputenc} 
\usepackage[T1]{fontenc}    
\usepackage{amsfonts}       
\usepackage{nicefrac}       
\usepackage{xcolor}         
\usepackage{wrapfig}
\usepackage{algorithm}
\usepackage{algorithmic}
\usepackage{bm}
\usepackage{tikz}
\usepackage{fontawesome}
\usepackage{amsmath}
\usepackage{amssymb}
\usepackage{bbm}
\usepackage{pifont}
\usepackage{fancybox}
\usepackage{tikz}
\usepackage{enumitem}
\usepackage{color, colortbl}
\usepackage{dsfont}
\usepackage{multirow} 
\usepackage[referable]{threeparttablex}
\usetikzlibrary{positioning}
\usetikzlibrary{arrows}
\usetikzlibrary{calc}
\usetikzlibrary{fit, shadows, shapes, backgrounds}
\usetikzlibrary{bayesnet}
\usepackage{url}
\usepackage{xr}
\usepackage{comment}
\usepackage[toc,page,header]{appendix}
\usepackage{minitoc}

\renewcommand \thepart{}
\renewcommand \partname{}

\usetikzlibrary{shapes,decorations,arrows,calc,arrows.meta,fit,positioning}
\tikzset{
    -Latex,auto,node distance =1 cm and 1 cm,semithick,
    state/.style ={ellipse, draw, minimum width = 0.7 cm},
    point/.style = {circle, draw, inner sep=0.04cm,fill,node contents={}},
    bidirected/.style={-Latex,dashed},
    el/.style = {inner sep=2pt, align=left, sloped},
   box/.style={rectangle,draw,node distance=1cm,text width=15em,text centered,rounded corners,minimum height=2em,thick},
    arrow/.style={draw,-latex',thick,dashed},
}
\definecolor{Gray}{gray}{0.9}
\definecolor{LightCyan}{rgb}{0.8,0.9,0.9}

\definecolor{newcolor}{rgb}{.8,.349,.1}
\newcolumntype{a}{>{\columncolor{blue!15}}c}

\renewlist{tablenotes}{enumerate}{1}
\makeatletter
\setlist[tablenotes]{label=\tnote{\alph*},ref=\alph*,itemsep=\z@,topsep=\z@skip,partopsep=\z@skip,parsep=\z@,itemindent=\z@,labelindent=\tabcolsep,labelsep=.1em,leftmargin=*,align=left,before={\scriptsize}}

\usepackage[utf8]{inputenc} 
\usepackage[T1]{fontenc}    
\definecolor{cvprblue}{rgb}{0.21,0.49,0.74}
\usepackage[pagebackref,breaklinks,colorlinks,allcolors=cvprblue]{hyperref}
\usepackage{url}            
\usepackage{booktabs}       
\usepackage{amsfonts}       
\usepackage{nicefrac}       
\usepackage{microtype}      
\usepackage{xcolor}         

\usepackage{colortbl} 

\newcommand{\gr}{\color{green!50!black}}

\usepackage{hyperref}

\usepackage{orcidlink}

\begin{document}

\title{Fourier Self-Supervision for Fine-Grained Generalized Category Discovery}
%

\author{Sarah Rastegar\inst{1}\orcidlink{0000-0002-4542-7388} \and
Mina Ghadimi Atigh\inst{1}\orcidlink{0009-0003-8377-270X} \and
Pascal Mettes\inst{1}\orcidlink{0000-0001-9275-5942}\and
Yuki M. Asano\inst{2}\orcidlink{0000-0002-8533-4020}\and
Cees G. M. Snoek\inst{1}\orcidlink{0000-0001-9092-1556}}

\authorrunning{S.~Rastegar et al.}

\institute{University of Amsterdam \and
University of Technology Nuremberg
}

\titlerunning{Fourier Self-Supervision}

\maketitle
\begin{abstract}
Generalized Category Discovery aims to recognize known categories while identifying novel ones within unlabeled data. Existing methods, typically based on self-supervision and contrastive learning, often struggle to capture fine-grained distinctions, relying on superficial visual cues rather than the intrinsic attributes humans use for categorization. We introduce Fourier Self-Supervision, that leverages the Fourier transform of images to enhance the discrimination of subtle differences and support the discovery of new categories. Our method employs a dual frequency filtering strategy: a low-pass filter first extracts broad, abstract attributes that capture high-level category information, while a high-pass filter emphasizes fine details such as edges and textures that are essential for fine-grained recognition. Each operates on a dedicated latent space, and their overlapping representations together yield a richer, more complete feature space. This dual-frequency approach not only refines feature extraction to identify novel categories, but also strengthens the model’s discriminative power in fine-grained category discovery. Experiments on multiple fine-grained datasets show that incorporating Fourier Self-Supervision outperforms state-of-the-art methods, even when the number of classes is unknown, demonstrating its effectiveness for Generalized Category Discovery. Our code is available at: \url{https://github.com/SarahRastegar/FourEx}.
\end{abstract}    

\label{sec:intro}
\begin{figure}[h]

\centering
\includegraphics[width=1\linewidth]{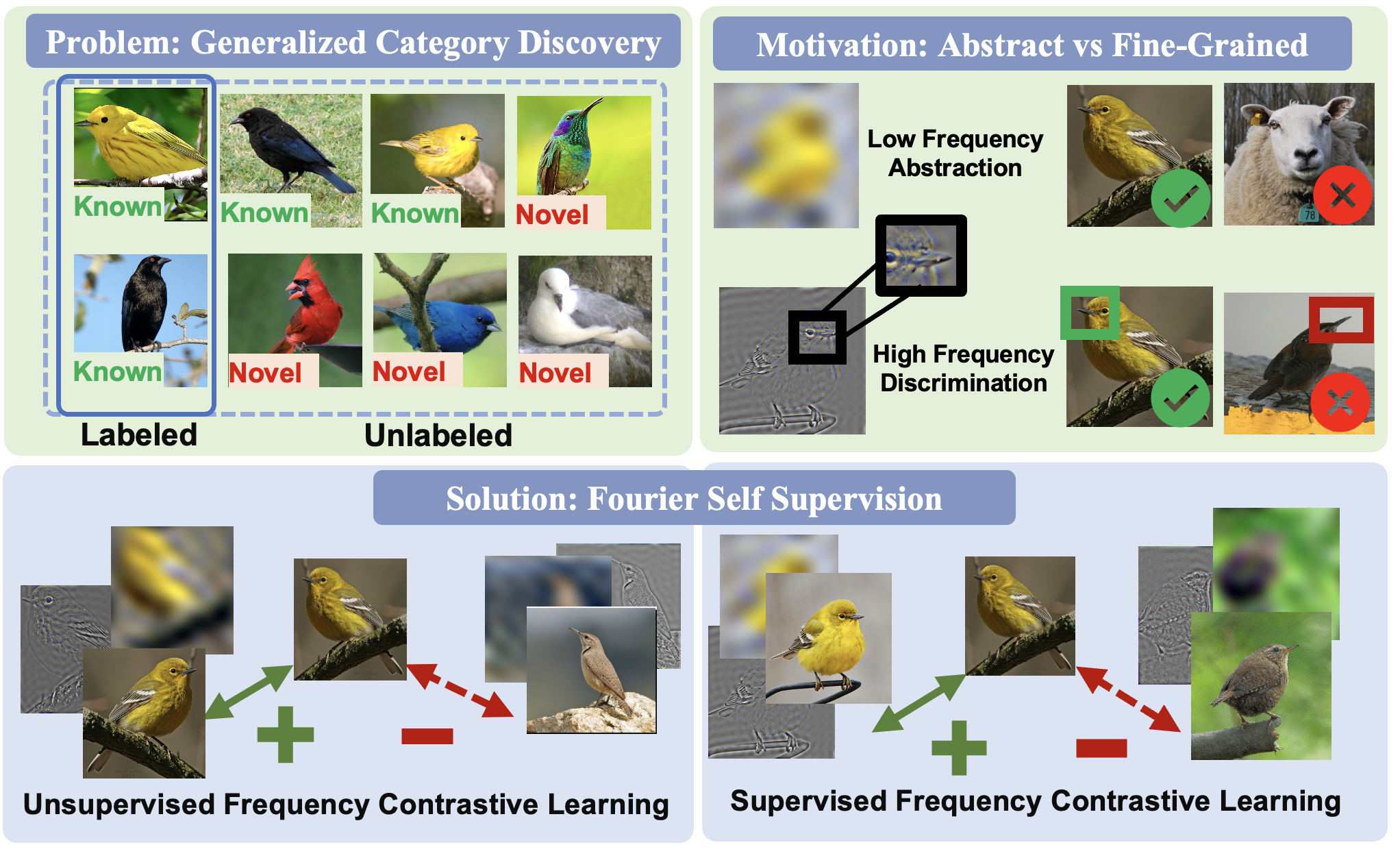}
\caption{\textbf{Overview of Our Approach.} \textit{(a) Problem}, We aim to cluster unknown, novel categories alongside known ones. \textit{(b) Motivation}, The blurred image of a yellow bird, reconstructed from low Fourier frequencies, provides a general sense of the category, facilitating broad generalizations. In contrast, high-frequency reconstructions reveal finer details, such as beaks and feathers. \textit{(c) Fourier Self-Supervision} harnesses the generalizability of low frequencies and the fine-grained discrimination of high frequencies in Fourier self-supervision to discover novel, fine-grained categories. }\label{fig:fig1_one}
\end{figure}

\section{Introduction}
\begin{figure*}[t!]
  \centering
  \includegraphics[width=1\linewidth]{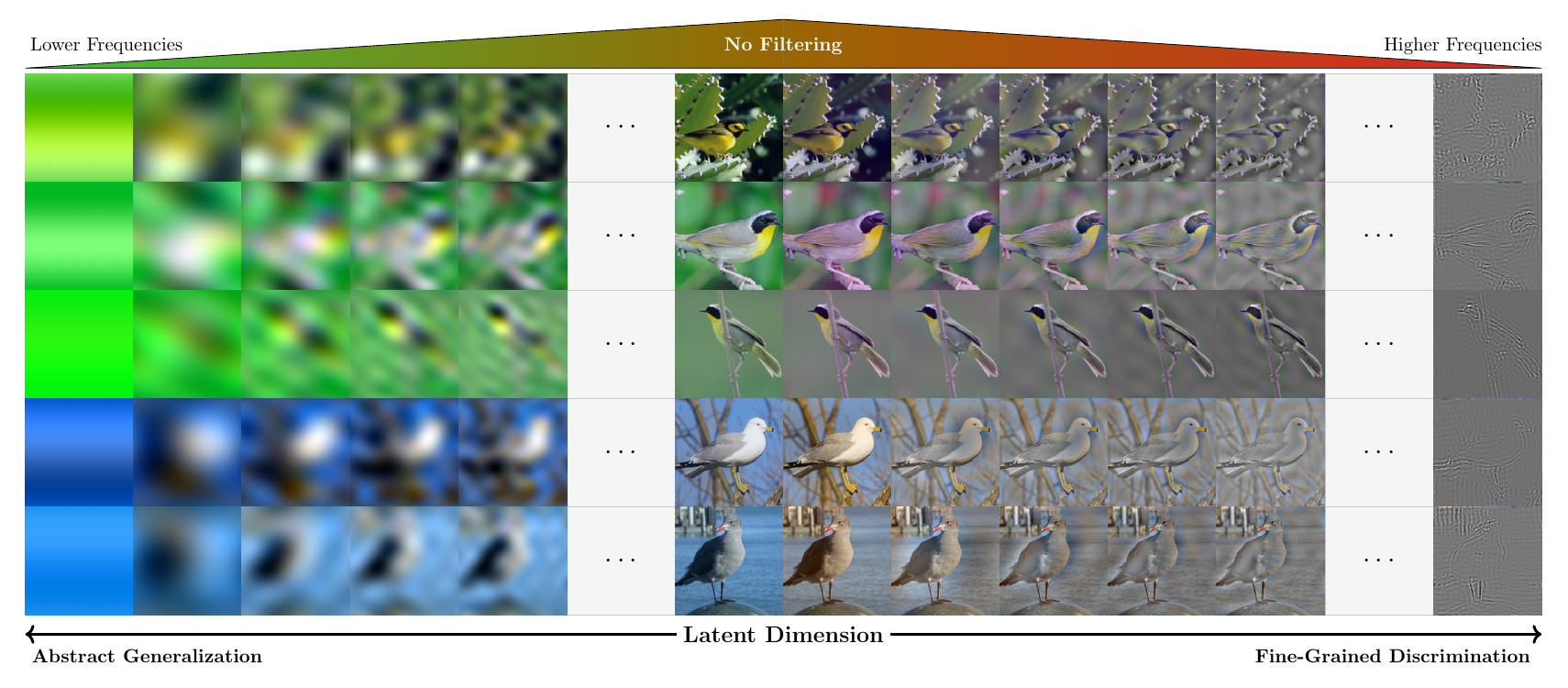}
\caption{\textbf{Motivation for Fourier Self-Supervision.} Low frequencies support broad category generalization (e.g., marine vs. forest birds), while high frequencies enable fine-grained recognition. By progressively integrating Fourier components and disentangling them in the latent space, our model builds hierarchical representations for both generalization and accuracy.} 
\label{fig:frame_motive}
\end{figure*}

Generalized Category Discovery (GCD) mitigates the inability of supervised learning to recognize novel categories, by enabling the recognition of both known and novel categories within unlabeled datasets \cite{cao2021open,vaze2022generalized,zhang2022promptcal, hao2024cipr,chiaroni2023parametric,wen2022simple, an2023generalized,rastegar2023learn,vaze2023clevr4}.
A prominent GCD strategy for addressing the challenge of unknown categories is self-supervision via contrastive learning \cite{jaiswal2020survey,zhai2019s4l,liu2021self,chen2020simple, li2021prototypical, noroozi2016unsupervised,caron2018deep, cao2021open, han2020automatically}, which utilizes data augmentation to produce different views of the same image \cite{he2020momentum, chen2020simple, caron2020unsupervised, caron2021emerging, oquab2024dinov}. Current literature mostly focuses on the coarse-grained setting, where categories have clear appearance differences. But as categories become more fine-grained and inter-class appearance differences start to shrink, we should still be able to figure out when a sample is part of a known set of labels, or is part of a new category. 
To tackle the fine-grained GCD challenge, we draw inspiration from Fourier analysis. Intuitively, an object’s categorization should remain consistent across the frequency spectrum of its Fourier transform. However, broader, more general categories tend to manifest in lower frequencies, while fine-grained distinctions emerge more clearly in higher frequencies (see \cref{fig:frame_motive}).
Building on this, we introduce a frequency-based self-supervision to enhance category discovery, particularly when classes exhibit strong visual similarity. The overall problem setup, motivation, and a high-level overview of our method are illustrated in \cref{fig:fig1_one}.

Incorporating Fourier self-supervision allows us to exploit the F-principle \cite{xu2019training}, which posits that overparameterized neural networks generalize well partly because they prioritize learning lower Fourier components of a function before gradually adjusting to higher frequencies. To extend generalizability to novel fine-grained categories, we aim for the model to infer unseen categories using the knowledge acquired from known ones. When categories adhere to an implicit hierarchical structure, observing a sibling category and its shared parent can help constrain the search space for novel categories. This implicit hierarchy, inferred through lower frequency components, allows the model to rapidly differentiate novel categories by focusing on their shared traits with known siblings, reducing both parent and sibling category confusion. For instance, in the left half of the latent dimension in \cref{fig:frame_motive}, lower frequencies provide a coarse-grained, general view of a category, allowing the model, for example, to differentiate whether a blurred image of a bird originates from a marine or forest environment. By progressively incorporating higher Fourier components, the model can more effectively distinguish fine-grained details as required for refined category discrimination. We propose an approach that assigns distinct parts of the latent space to capture these frequencies progressively, enabling the model to generalize to novel categories by learning an implicit hierarchy from low-frequency features, while high frequencies drive fine-grained differentiation. 


\noindent Our contributions are as follows:
\begin{itemize}
\item We apply low-pass filtering in the Fourier domain to help the model capture broad, abstract characteristics of images, enabling it to infer an implicit hierarchical structure that improves generalization to unseen categories.
\item We employ high-pass filtering to focus the model’s attention on fine details and enhance its ability to distinguish subtle differences for fine-grained classification.
\item Through empirical evaluation, we show that our method outperforms state-of-the-art approaches on fine-grained generalized category discovery tasks, while also achieving competitive performance on coarse-grained datasets.
\end{itemize}

\section{Related Works}

\noindent\textbf{Generalized Category Discovery.}
The task of Generalized Category Discovery was formalized by Vaze~\etal~\cite{vaze2022generalized} and Cao~\etal~\cite{cao2021open}. GCD lies at the intersection of supervised and unsupervised learning. It leverages a small amount of labeled data along with a larger set of unlabeled data, where the unlabeled data can contain both known and novel categories. This makes GCD a special case of self-supervised learning \cite{ouali2020overview, yang2022survey, rebuffi2020semi, oliver2018realistic, chapelle2009semi}. There are two prominent approaches for GCD: (i) \textit{Prototype-based Methods:} These approaches leverage a set of pre-defined prototypes as reference points to guide category discovery in the unlabeled data. This can be achieved through techniques like nearest neighbor search or learning a distance metric that effectively separates known and unknown categories, \cite{hao2024cipr, chiaroni2023parametric, wen2022simple, an2023generalized}. 
(ii) \textit{Similarity-driven Clustering:} This approach focuses on exploiting local similarities within the unlabeled data to form initial category clusters. This can be done through techniques like k-nearest neighbors or using mean-teacher frameworks to mitigate the issue of noisy pseudo-labels generated from similar information
\cite{pu2023dynamic,zhang2022promptcal, hao2024cipr, chiaroni2023parametric, rastegar2023learn, banerjee2024amend, otholt2024guided,vaze2023clevr4,zhang2022promptcal, wen2022simple}. Nonetheless, Contrastive learning methods often struggle with fine-grained GCD due to aggressive augmentations overshadowing subtle category differences \cite{cole2022does}. Our work addresses this by leveraging the informative nature of high frequencies for fine-grained discrimination, while utilizing the lower frequencies for category discovery, enabling better handling of nuanced visual details. Several recent works explore alternative approaches for GCD. Hierarchical approaches proposed by Otholt~\etal~\cite{otholt2024guided} and Banerjee~\etal~\cite{banerjee2024amend} leverage neighborhood structures for refined category delineation. Choi~\etal~\cite{choi2024contrastive} focus on robustness to noise by employing the mean shift algorithm for category discovery. Additionally, Wang~\etal\cite{wang2024sptnet} propose a spatial prompt tuning method that incorporates spatial information from image data to focus better on specific object parts. Differing from all these works, our method leverages weak supervision by focusing on high frequencies to distinguish fine-grained categories from each other while focusing on low frequencies to generalize categorization to novel ones.

\noindent\textbf{Fine Grained Generalized Category Discovery.}
One of the main challenges in generalized category discovery arises when categories are so fine-grained that traditional clustering in feature space struggles to accurately identify novel categories. To address this, prior work has taken several approaches: Fei~\etal~\cite{fei2022xcon} partition data into expert sub-datasets by applying k-means clustering on self-supervised representations, while Rastegar~\etal~\cite{rastegar2023learn} proposed implicit category trees that enable hierarchical self-coding, preserving category similarity across multiple levels. Liu~\etal~\cite{liu2025generalized} introduce class-wise distribution regularization to alleviate the bias towards known categories. Meanwhile, Rastegar~\etal~\cite{RastegarECCV2024} introduced the concept of self-expertise, which leverages abstract pseudolabels to facilitate distinction between samples. Many recent methods~\cite{dai2025adaptive,he2025seal,peng2025mos,yang2024learning} exploit subtle sample-level variations to capture fine-grained details. In contrast, our approach selectively leverages frequency information, it emphasizes high-frequency components to enhance fine-grained discrimination, while relying on low frequencies to improve generalization to novel categories.

\noindent\textbf{Fourier Transform based Image Augmentation.}
Fourier image augmentation has gained attention in computer vision due to its potential to enhance the robustness and generalizability of deep learning models. It leverages the Fourier transform to modify image data in the frequency domain, providing unique advantages over traditional spatial domain augmentations. Yang~\etal~\cite{yang2020fda} proposed Fourier Domain Adaptation, which applies the Fourier transform to both source and target domain images to align their frequency distributions. This improves classification accuracy considerably in domain adaptation scenarios by focusing on the low-frequency components that capture essential structural information while ignoring high-frequency noise. Sun~\etal~\cite{sun2022spectral} introduced FourierMix, a data augmentation technique that blends images in the frequency domain. This method creates new training examples by mixing the amplitude and phase spectra of different images, leading to improved generalization performance on various benchmark datasets. Xu~\etal~\cite{xu2021fourier} used Fourier magnitude swaps between different samples to improve domain generalization. In this work, we take inspiration from such methods and we introduce a self-supervised approach that leverages these properties of Fourier transforms to enable concept discovery in fine-grained settings, where details matter.
\section{Preliminaries}
\noindent\textbf{Contrastive Learning for GCD.}
In a mini-batch with size $B$, let $\mathbf{x}_i$ and $\mathbf{x}_i'$ denote two distinct augmentations of the same image. The unsupervised contrastive loss $\mathcal{L}_i^u$ and its supervised counterpart $\mathcal{L}_i^s$ are formally described as  \cite{vaze2022generalized}:
\begin{equation}
    \label{eq:unsupd}
    \mathcal{L}_i^u{=}-\log{}\frac{e^{\mathbf{z}_i.\mathbf{z}_i'/\tau}}{\sum_n\mathds{1}_{[n\neq i]}e^{\mathbf{z}_i.\mathbf{z}_n'/\tau}},
\end{equation}
\begin{equation}
    \label{eq:supd}
    \mathcal{L}_i^s{=}-\frac{1}{|\mathcal{N}(i)|}\sum_{q\in\mathcal{N}(i)}\log{}\frac{e^{\mathbf{z}_i.\mathbf{z}_q/\tau}}{\sum_n\mathds{1}_{[n\neq i]}e^{\mathbf{z}_i.\mathbf{z}_n/\tau}}.
\end{equation}
In the proposed formulation, each image $\mathbf{x}_i$ is associated with an embedding $\mathbf{z}_i$, where $\mathbf{z}_i.\mathbf{z}_i'$ represents a similarity metric that may encompass cosine similarity, Euclidean distance, or other relevant measures. The indicator function $\mathds{1}{[n\neq i]}$ is defined such that it equals 1 if and only if $n \neq i$, and $\tau$ is a temperature hyperparameter. Additionally, the unsupervised loss can be defined selectively for specific segments of the embedding vector. 
where $\mathcal{N}(i)$ denotes the collection of images within the same minibatch that are assigned the identical label as image $\mathbf{x}_i$. 
Finally, for generalized category discovery, Vaze~\etal~\cite{vaze2022generalized} consider the total loss as: 
\begin{equation}
    \label{eq:gcd}
    \mathcal{L}^g{=}(1{-}\lambda)\sum_{i\in B}\mathcal{L}_i^u{+}\lambda\sum_{i\in B_{\mathcal{L}}}\mathcal{L}_i^s.
\end{equation}
where $B_{\mathcal{L}}$ is the labeled portion of minibatch $B$ and $\lambda$ is a mixing hyperparameter.

\noindent\textbf{Fourier Transform.}
For an image $\mathbf{x}_i{=}f(h,y)$, its Fourier transformation $\mathcal{F}(u,v)$ and its reconstruction by using inverse transformation $\mathcal{F}^{-1}$is formulated as:
\begin{equation}
    \label{eq:fourier}
    \mathcal{F}{(u,v)}{=}\sum_{h{=}0}^{H-1}\sum_{w{=}0}^{W-1}f(h,w)e^{-2\pi{}i (\frac{h}{H}u{+}\frac{w}{W}v)},
\end{equation}
\begin{equation}
    \label{eq:infourier}
    f{(h,w)}{=}\frac{1}{WH}\sum_{u{=}0}^{H-1}\sum_{v{=}0}^{W-1}\mathcal{F}(u,v)e^{2\pi{}i (\frac{u}{H}h{+}\frac{v}{W}w)}.
\end{equation}
The efficient computation of both the Fourier transform and its inverse is facilitated by the Fast Fourier Transform (FFT) algorithm, as detailed by~\cite{nussbaumer1982fast}. A low-pass filter (LPF), characterized by its transfer function $H_{LP}(u,v)$, incorporates a cutoff frequency denoted by $f_{l}$. This function defines the operational boundaries of the LPF as $H_{LP}(u,v){=}\mathds{1}(|u|{<}f_{l},|v|{<}f_{l})$.
 Correspondingly, a high pass filter (HPF) is formulated as $H_{HP}(u,v){=}\mathbf{1}-H_{LP}(u,v)$ in which $\mathbf{1}$ represents the matrix of ones. The operation involves the convolution of the image with the respective filter kernel to implement either a low-pass or high-pass filter. This is efficiently achieved in the frequency domain by the element-wise multiplication of the image's Fourier transform with the filter's transfer function.
 \begin{wrapfigure}{r}{0.5\textwidth}
  \centering
  \includegraphics[width=\linewidth]{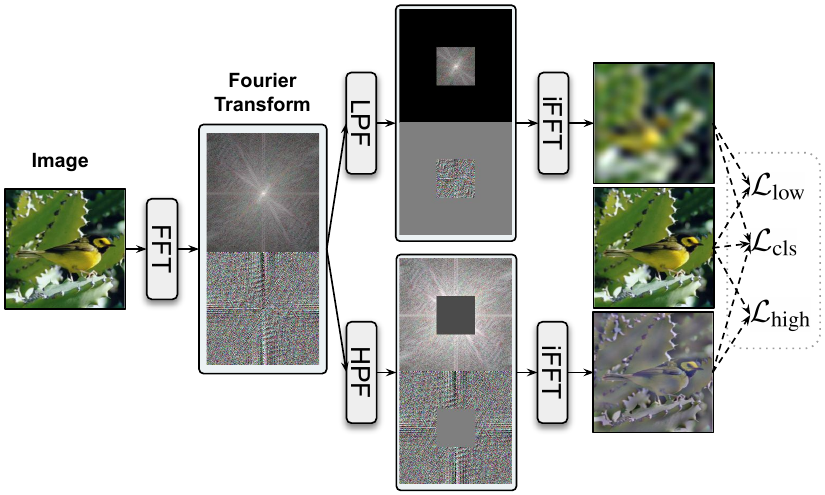}
\caption{\textbf{Fourier Self-Supervision:} An image undergoes a Fourier transform to extract its magnitude and phase. Both low-pass and high-pass filters are applied in parallel in the frequency domain. The inverse Fourier transform then reconstructs the frequency-filtered images, which are used as positive views for the original image in contrastive learning. Finally, all three views are utilized to predict the class of labeled images.} 
\label{fig:pipe}
\vspace{-4.5em}
\end{wrapfigure}
\section{Fourier Self-Supervision}

Inspired by the principles of Fourier optics \cite{goodman2005introduction} and recognizing that the human eye lens performs a natural Fourier transform, we propose that object classification can be discerned across multiple frequency bands. For this purpose, we define two specific cutoff frequencies, denoted as 
$f_{l}$ and $f_{h}$. As shown in \cref{fig:pipe}, to isolate lower frequencies, we employ a low-pass filter (LPF). Conversely, to eliminate lower frequency components, a high pass filter (HPF). In the subsequent sections, we provide a detailed description of how each filter is implemented.

\noindent\textbf{Extraction of Critical Frequencies.} 
~To effectively implement both low-pass and high-pass filtering, we define a threshold frequency to serve as an upper bound, ensuring that the model is not influenced by very high frequencies that typically contain noise, which can obscure informative patterns. We introduce a high threshold value, $T_h$, to cap these frequencies. For low-pass filtering, if the cutoff frequency is too high, the image retains excessive detail, offering little additional learning benefit beyond the original image. To make this threshold dataset-dependent rather than arbitrary, we determine the maximum threshold frequency that maintains a signal-to-noise ratio (SNR) of $15$ dB, aligning with current standards for acceptable subjective image and video quality~\cite{winkler2005digital,poynton2012digital}. 
To achieve category-based representations, we compute the high-frequency cutoff threshold across all categories in the dataset and utilize the average value to ensure a balanced and dataset-dependent approach. The high-frequency cutoff is determined by maintaining a signal-to-noise ratio (SNR) that aligns with a standard of $15$ dB, ensuring both informative features and noise suppression. To compute the SNR for different cutoff frequencies, we first represent the Fourier transform of an image as $X$, and calculate the SNR as  $SNR(f_c){=}10\log(\frac{\sum_{f\leq f_c}|X(f)|^2}{\sum_{f> f_c}|X(f)|^2})$,
where $|.|^2$ is the average magnitude of a signal and 
$f_c$ denotes the cutoff frequency. By averaging this threshold across all categories, we establish a robust, dataset-dependent criterion for frequency filtering, effectively balancing generalization and fine-grained discrimination. We then select an SNR of $32$, equivalent to $15$ dB, as this is widely recognized as the standard for acceptable signal quality. To ensure robustness, we also conduct experiments with different SNR values, as detailed in the appendix. Using this approach, we determine the low-pass cutoff frequency threshold $T_l$ and its corresponding high-pass cutoff frequency threshold $T_h$. These thresholds are subsequently employed in the later stages of our method to guide both low-frequency generalization and high-frequency discrimination effectively.
\begin{wrapfigure}{l}{0.5\textwidth}
\vspace{-1em}
  \centering
  \includegraphics[width=\linewidth]{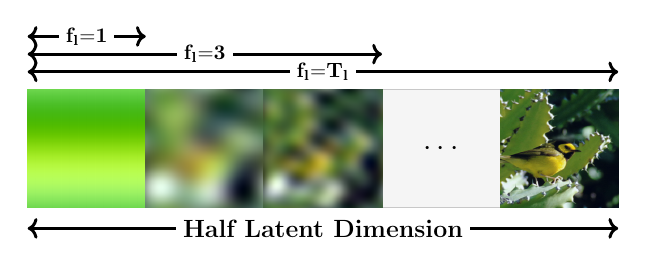}
 \vspace{-1.5em}
\caption{\textbf{Low-Frequency Contrastive Learning.} Each image undergoes low-pass and high-pass filtering at a random cut-off frequency, denoted as $f_l$. Then the ratio $\frac{f_l}{T_l}$ is calculated, where $T_l$ represents the maximum cutoff frequency. This ratio determines the fraction of the latent dimension dedicated to contrasting the filtered image against its original counterpart. In this example, a cutoff frequency of $f_l{=}1$, allocating only $\frac{1}{T_l}$ of its leftmost latent dimension for contrastive learning.}
\label{fig:lowcon}
\vspace{-1em}
\end{wrapfigure}

\noindent\textbf{Low-Frequency Contrastive Learning.}
Human perception remains capable of discerning image categories despite the absence of high-frequency details, which are typically associated with noise or subtle information. As frequency components are progressively omitted, there is a loss of categorical information; for example, in \cref{fig:lowcon}, initial frequency bands allow only the recognition of basic attributes like color or the identification of an object as a bird. To leverage this abstracted information without compromising overall representational quality, we select a terminal frequency band $T_{\text{l}}$ with an SNR of $15$ dB or more, beyond which differences are imperceptible to the human eye. Utilizing $\frac{f_{\text{l}}}{T_{\text{l}}}$ of the dimensionality, we apply contrastive learning to facilitate generalized category discovery through embeddings. Considering an image $\mathbf{x}_i$, its low-frequency reconstruction which is denoted by $\mathbf{l}_i$ is $\mathbf{l}_i{=}\mathcal{F}^{-1}(H_{LP}\odot\mathcal{F}(\mathbf{x}_i))$.
After the reconstruction phase, the image is employed as input for both supervised and unsupervised contrastive learning, specifically targeting a fraction $\frac{f_{l}}{T_l}$ of the latent dimension $D$. We define this fraction as $d {=} \frac{f_{l}}{T}D$ and incorporate it into \cref{eq:gcd} to compute the low-frequency contrastive loss:
\begin{equation}
    \label{eq:low}
    \mathcal{L}_{\text{low}}{=}(1{-}\lambda)\sum_{i\in B}\mathcal{L}_{i|d}^u{+}\lambda\sum_{i\in B_{\mathcal{L}}}\mathcal{L}_{i|d}^s.
\end{equation}
The loss $\mathcal{L}_{i|d}^u$ is defined based on the subset of its first $d$ leftmost latent dimension. 

\begin{wrapfigure}{r}{0.5\textwidth}

\vspace{-1em}
  \centering
  \includegraphics[width=\linewidth]{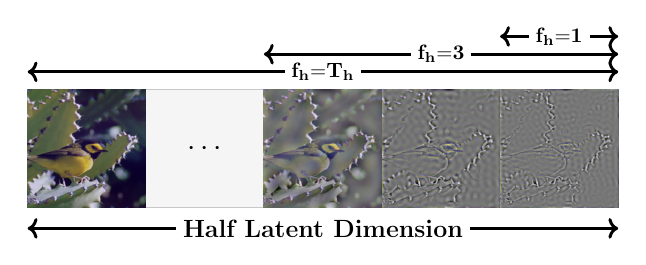}
\vspace{-1.5em}
\caption{\textbf{High-Frequency Contrastive Learning.} Each image undergoes high-pass filtering at a random cutoff frequency, denoted as $f_h$. Then the ratio $\frac{f_h}{T_h}$ is calculated, where $T_h$ represents the maximum cutoff frequency. This ratio determines the fraction of the latent dimension dedicated to contrasting the filtered image against its original counterpart. In this example, the square corresponding to $1$ is filtered at a cutoff frequency of $T_h-1$, allocating only $\frac{1}{T_h}$ of its rightmost latent dimension for contrastive learning.}
\label{fig:highcon}
\vspace{-1em}

\end{wrapfigure}

\noindent\textbf{High-Frequency Contrastive Learning.}
In this section, we explore high-frequency contrastive learning to enhance model sensitivity to fine-grained image details crucial for differentiating categories in fine-grained datasets. We select a random cutoff frequency $f_h{<}T_h$ to focus on these critical details. As depicted in  \cref{fig:highcon}, increasing $f_h$ beyond a certain threshold introduces noise and reduces informativeness. To optimize the use of these nuanced details for category distinction without degrading the quality of representation, we employ a strategy analogous to low-frequency contrastive learning. Here, we initiate from the rightmost latent representation, unlike the leftmost starting point used in low-frequency settings. Consider a frequency band $T_h$, beyond which category distinction becomes imperceptible to humans. We allocate the last $1{-}\frac{f_h}{T_h}$ of the dimension for contrastive learning. Considering an image $\mathbf{x}_i$, its high-frequency reconstruction is denoted by $\mathbf{h}_i$ which can be calculated as $\mathbf{h}_i{=}\mathcal{F}^{-1}(H_{HP}\odot\mathcal{F}(\mathbf{x}_i))$
Similar to the previous section, we integrate $\mathbf{h}_i$ as an alternative view of $\mathbf{x}_i$. For contrastive learning, we employ the last $d{=}\frac{f_h}{T}D$ dimensions, denoted as ${-}d$. This substitution is utilized in \cref{eq:gcd} to compute the high-frequency contrastive loss.
\begin{equation}
    \label{eq:high}
    \mathcal{L}_{\text{high}}{=}(1{-}\lambda)\sum_{i\in B}\mathcal{L}_{i|{-}d}^u{+}\lambda\sum_{i\in B_{\mathcal{L}}}\mathcal{L}_{i|{-}d}^s.
\end{equation}
In the Appendix, we show that fine details reside only in high-frequency modes; dropping them discards the information needed to distinguish those details. In practice, rather than allocating the entire 
$D$ latent dimension to both low and high-frequency components, we divide it such that the left half is dedicated to the low-frequencies, while the right half is assigned to the high-frequencies.

\noindent\textbf{Frequency Augmentation Classification.}
In addition to employing contrastive learning, our approach also acknowledges that the low-frequency reconstruction $\mathbf{l}_i$, high-frequency reconstruction $\mathbf{h}_i$, and the original sample $\mathbf{x}_i$ share identical ground truth labels. Consequently, we incorporate a binary cross-entropy loss function to facilitate the label classification from these instances. This enables a unified treatment of label consistency across different representation modalities within our framework.
\begin{equation}
    \label{eq:cls}
    \mathcal{L}_{\text{cls}}{=}\mathcal{L}_{\text{BCE}}(p_{h_i},y_i){+}\mathcal{L}_{\text{BCE}}(p_{l_i},y_i){+}\mathcal{L}_{\text{BCE}}(p_{x_i},y_i).
\end{equation}
In this formulation, $p_{h_i}$ denotes the labels predicted by the model for $\mathbf{h}_i$. Summing up, the total loss is computed as the aggregation of individual losses together with the baseline model's loss, thus we have:
\begin{equation}
    \label{eq:cls}
    \mathcal{L}_{\text{total}}{=}\mathcal{L}_{\text{base}}+\alpha_{\text{low}}\mathcal{L}_{\text{low}}+\alpha_{\text{high}}\mathcal{L}_{\text{high}}+\alpha_{\text{cls}}\mathcal{L}_{\text{cls}}.
\end{equation}

\section{Experiments}
\label{sec:experiment}
\subsection{Experimental Setup}

\textbf{Datasets.}
We evaluate our method on three fine-grained datasets: CUB-200~\cite{wah_branson_welinder_perona_belongie_2011}, FGVC-Aircraft~\cite{maji2013fine}, and Stanford-Cars~\cite{krause20133d}. Further, we include the challenging Herbarium dataset~\cite{tan2019herbarium}, which is both fine-grained and long-tailed, demonstrating that our approach remains effective even under severe class imbalance. We also report results on Oxford-IIIT Pet~\cite{parkhi2012cats}, a small-scale fine-grained dataset with limited samples per category, making it particularly challenging.
In the Appendix, we further evaluate our approach on coarse-grained datasets, CIFAR-10~\cite{krizhevsky2009learning}, CIFAR-100~\cite{krizhevsky2009learning}, and ImageNet-100~\cite{deng2009imagenet} to demonstrate its adaptability to both fine-grained and coarse-grained classification. Detailed dataset statistics and train/test split information are also included in the Appendix.

\noindent\textbf{Implementation Details.}
In our experiments, we adopted the data partitioning strategy by Vaze~\etal~\cite{vaze2022generalized}, wherein half the categories for each dataset are recognized as known.  
From these known categories, half of the samples form the labeled set, while the residual data from known categories, along with all samples from novel categories, are allocated to the unlabeled set. Following the approach by Vaze~\etal~\cite{vaze2022generalized}, our primary model architecture is the ViT-B/16, utilizing either a DINOv1~\cite{caron2021emerging} pre-training on the unlabeled ImageNet1K~\cite{krizhevsky2017imagenet} dataset or a DINOv2~\cite{oquab2024dinov} pre-training on the LVD-142M dataset. We apply Fourier Self-Supervision to baselines SimGCD~\cite{wen2022simple} and SelEx~\cite{RastegarECCV2024}, resulting in two variants we call \textit{FourSim} and \textit{FourEx}, respectively. For FourSim, we froze the initial 11 blocks of the ViT-B/16 and fine-tuned the last block, while for FourEx, we finetuned last three blocks. In the Appendix, we show that independent of the number of frozen blocks, our method improves over both baselines. Implementation details for augmenting other backbones with Fourier Self-Supervision are provided in the Appendix.
\begin{table*}[ht]
      \caption{\textbf{Comparison with state-of-the-art for fine-grained image classification.} Bold and underlined numbers indicate the best and second-best accuracies, respectively. Our method is well-suited for fine-grained datasets, profits from stronger backbones, and has strong performance for all three experimental settings (\textit{All}, \textit{Known}, and \textit{Novel}).}
  \centering
  \begin{threeparttable}
  \resizebox{1\linewidth}{!}{
\begin{tabular}{cllaccaccaccacc}
\toprule
&&&\multicolumn{3}{c}{\textbf{CUB-200}}& \multicolumn{3}{c}{\textbf{FGVC-Aircraft}}&\multicolumn{3}{c}{\textbf{Stanford-Cars}}&\multicolumn{3}{c}{\textbf{Average}}\\ \cmidrule(lr){4-6} \cmidrule(lr){7-9} \cmidrule(lr){10-12} \cmidrule(lr){13-15}&\textbf{Method}&\textbf{Venue}&All&Known&Novel&All&Known&Novel&All&Known&Novel&All& Known&Novel\\
\midrule
\multirow{14}{*}{\rotatebox{90}{DINOv1}}&GCD \cite{vaze2022generalized}	&{\footnotesize\textcolor{gray}{\textit{CVPR22}}}&51.3&	56.6&	48.7&	45.0	&41.1&46.9&39.0&57.6&29.9&45.1&51.8&41.8\\
&PromptCAL~\cite{zhang2022promptcal}&{\footnotesize\textcolor{gray}{\textit{CVPR23}}}	&62.9&64.4&	62.1&	52.2&	52.2&	52.3&50.2&70.1&	40.6&55.1&62.2&51.7\\
&$\mu$GCD~\cite{vaze2023clevr4}&{\footnotesize\textcolor{gray}{\textit{NeurIPS23}}}&65.7&68.0&64.6& 53.8&55.4&53.0&56.5&68.1 &50.9&58.7&63.8&56.2\\
&InfoSieve~\cite{rastegar2023learn}  &{\footnotesize\textcolor{gray}{\textit{NeurIPS23}}}               & 69.4 &\bf{77.9} &65.2 
 &56.3 &63.7 &52.5&  55.7 &74.8 &46.4&60.5&72.1&54.7\\
&SPTNet~\cite{wang2024sptnet}&{\footnotesize\textcolor{gray}{\textit{ICLR24}}}&65.8&68.8&65.1&59.3&61.8&58.1&59.0&79.2&49.3&61.4&69.9&57.5\\
&CMS~\cite{choi2024contrastive}&{\footnotesize\textcolor{gray}{\textit{CVPR24}}}&68.2&76.5&64.0&56.0&63.4&52.3&56.9&76.1 &47.6&60.4&72.0&54.6\\
&TRAILER~\cite{xiao2024targeted}&{\footnotesize\textcolor{gray}{\textit{CVPR24}}}&65.1&71.3&54.5&61.9&62.6& 50.5& 55.4& 71.7& 47.6&61.0&64.5&54.7\\  
&LegoGCD~\cite{cao2024solving}&{\footnotesize\textcolor{gray}{\textit{CVPR24}}}&63.8&71.9&59.8& 55.0& 61.5& 51.7 &57.3& 75.7 &48.4&58.7&69.7&53.3\\
&SelEx~\cite{RastegarECCV2024} &{\footnotesize\textcolor{gray}{\textit{ECCV24}}}&73.6 &75.3 &72.8&57.1 &64.7&53.3& 58.5 &75.6&50.3&63.0&71.9&58.8\\
&FlipClass~\cite{lin2024flipped}&{\footnotesize\textcolor{gray}{\textit{NeurIPS24}}}&71.3&71.3&71.3&59.3&\underline{66.9}& 55.4&63.1&\bf{81.7}&53.8&64.6&73.3&60.2\\
&DebGCD~\cite{liu2025debgcd}&{\footnotesize\textcolor{gray}{\textit{ICLR25}}}&66.3 &71.8 &63.5&  61.7& 63.9& 60.6&\bf{65.3} &\underline{81.6}& \underline{57.4}& 64.4&72.4&60.5\\
&APL~\cite{dai2025adaptive}&{\footnotesize\textcolor{gray}{\textit{CVPR25}}}&68.5 &73.1& 66.2&  60.9 &63.5 &59.6 &62.3 &80.7 &53.4&63.9& 72.4& 59.7\\
&MOS~\cite{peng2025mos}&{\footnotesize\textcolor{gray}{\textit{CVPR25}}}&69.6 &72.3& 68.2&  61.1& \underline{66.9}& 58.2&64.6 &80.9& 56.7& \underline{65.1} &\underline{73.4} &61.0\\
&AllGCD~\cite{cao2025allgcd}&{\footnotesize\textcolor{gray}{\textit{ICCV25}}}&68.4& 75.1& 65.1 & 57.4& 62.4& 54.9&60.5 &77.0& 52.5& 62.1& 71.5&57.5\\ 
&SEAL~\cite{he2025seal}&{\footnotesize\textcolor{gray}{\textit{NeurIPS25}}}&66.2 &72.1& 63.2&  \underline{62.0}& 65.3& 60.4&\bf{65.3} &79.3& \bf{58.5}& 64.5&72.2&60.7\\
\cmidrule{1-15}
&SimGCD~\cite{wen2022simple}&{\footnotesize\textcolor{gray}{\textit{ICCV23}}}&60.3&65.6&57.7&	54.2&	59.1&	51.8&53.8&71.9&45.0 &56.1&65.5&51.5\\
\rowcolor{gray!25}&\textbf{FourSim (Ours)} &&70.3&75.5&67.7& 56.1 &63.1 &52.6&56.6 &78.9 &45.8&61.0&72.5&55.4 \\
&SelEx$^\dag$~\cite{RastegarECCV2024} &{\footnotesize\textcolor{gray}{\textit{ECCV24}}}&76.4 &72.4 &78.4 &61.2 &67.7& 58.0 &56.9 &76.9 &47.3 &64.8&72.3&61.2\\
\rowcolor{gray!25}&\textbf{FourEx (Ours)} &&\bf{80.2}& 75.1& \bf{82.7}&\bf{65.9} &\bf{67.9}&\bf{64.9}&59.2 &76.9& 50.6&\bf{68.4}&73.3&\bf{66.1} \\
&Avg $\Delta$&&\textcolor{green!50!black}{+6.9}& \textcolor{green!50!black}{+6.3}&\textcolor{green!50!black}{+7.2}&\textcolor{green!50!black}{+3.3}&\textcolor{green!50!black}{+2.1}&\textcolor{green!50!black}{+3.9}&\textcolor{green!50!black}{+2.6}&\textcolor{green!50!black}{+3.5}&\textcolor{green!50!black}{+2.1}&\textcolor{green!50!black}{+4.3}&\textcolor{green!50!black}{+4.0}&\textcolor{green!50!black}{+4.4} \\
\midrule

\multirow{5}{*}{\rotatebox{90}{DINOv2}}
&GCD$^*$\cite{vaze2022generalized}&{\footnotesize\textcolor{gray}{\textit{CVPR22}}}&71.9& 71.2& 72.3&55.4 &47.9& 59.2& 65.7&67.8&64.7&64.3&62.3&65.4\\
&SimGCD$^*$~\cite{wen2022simple}&{\footnotesize\textcolor{gray}{\textit{ICCV23}}}&71.5&78.1&68.3& 63.9& 69.9 &60.9&71.5&81.9&66.6&69.0&76.6&65.3\\

&$\mu$GCD$^*$~\cite{vaze2023clevr4}&{\footnotesize\textcolor{gray}{\textit{NeurIPS23}}}&74.0&75.9&73.1&66.3&68.7&65.1&76.1& 91.0& 68.9&72.1&78.5&69.0\\
&FlipClass~\cite{lin2024flipped}&{\footnotesize\textcolor{gray}{\textit{NeurIPS24}}}&79.3&80.7&78.5&71.1&75.1& 69.1&78.0&88.0&73.2&76.1&81.3&73.6\\
&DebGCD~\cite{liu2025debgcd}&{\footnotesize\textcolor{gray}{\textit{ICLR25}}}&77.5 &80.8 &75.8 & 71.9& 76.0& 69.8&75.4 &87.7& 69.5& 74.9&81.5&71.7\\
&SEAL~\cite{he2025seal}&{\footnotesize\textcolor{gray}{\textit{NeurIPS25}}}&76.7& 78.3& 75.9&  74.6& 73.2& 75.3&77.7 &88.7 &72.4& 76.3&80.1&74.5\\
\cmidrule{1-15}
&SelEx~\cite{RastegarECCV2024}            &{\footnotesize\textcolor{gray}{\textit{ECCV24}}}&\underline{87.4}&\underline{85.1}&\underline{88.5}&  \underline{79.8}& \underline{82.3} &\underline{78.6}&  \bf{82.2} &\underline{93.7}&\bf{76.7}&\bf{83.1}&\underline{87.0}&\bf{81.3}\\
\rowcolor{gray!25}&\textbf{FourEx (Ours)} &         & \bf{87.8} &\bf{86.3} &\bf{88.6}
 &  \bf{81.5} &\bf{84.7} &\bf{80.0}
&  \underline{80.1} &\bf{94.3} &73.2&\bf{83.1}&\bf{88.4}&\underline{80.6}\\
&Avg $\Delta$&&\textcolor{green!50!black}{+0.4}& \textcolor{green!50!black}{+1.2}&\textcolor{green!50!black}{+0.1}&\textcolor{green!50!black}{+1.7}&\textcolor{green!50!black}{+2.4}&\textcolor{green!50!black}{+1.4}&\textcolor{red!50!black}{-2.1}&\textcolor{green!50!black}{+0.6}& \textcolor{red!50!black}{-3.5}&+0.0&\textcolor{green!50!black}{+1.4}&\textcolor{red!50!black}{-0.7} \\
\bottomrule
\end{tabular}
}
\begin{tablenotes}
      \item[]$^*$ Reported from \cite{vaze2023clevr4}. $^\dag$ Results for fine-tuning the last three blocks.
    \end{tablenotes}
    \end{threeparttable}

  \label{tab:expcub}
\end{table*}

Before training, we measure the signal-to-noise ratio (SNR) across a range of cutoff frequencies and select the cutoff that produces an SNR of 15 dB, a standard commonly used in video quality assessment. This procedure only requires applying the Fourier transform to a batch of training images to estimate the SNR for different cutoff frequencies. The corresponding cutoff values for different datasets and SNR levels are provided in the Appendix. In practice, we observe that the granularity of a dataset largely determines the resulting SNR values. Furthermore, to mitigate Gibbs ringing artifacts near image edges, we apply a Gaussian blur with a kernel size of 5 to each image before filtering.

\subsection{Comparison with State-of-the-Art}

\noindent \textbf{Fine-grained image classification.}
In \cref{tab:expcub}, we evaluate the performance of FourEx on three fine-grained datasets. The results show that our method excels at fine-grained categorization, outperforming existing approaches in both the \textit{All} and \textit{Novel} category settings for both DINOv1 and DINOv2 backbones. This improvement arises from the model’s ability to separate subtle high-frequency details from broader low-frequency patterns, enabling more precise recognition of fine-grained categories. Both \textit{FourSim} and \textit{FourEx}, consistently improve over their respective baselines when using a DINOv1 backbone. A similar trend is observed with DINOv2, with the exception of the Stanford Cars dataset. This dataset is comparatively less fine-grained than the others, as the differences between cars are more pronounced. Consequently, the improvements of our method are more evident on finer-grained datasets.

As shown in \cref{tab:pets}, our model also performs consistently well on the Oxford-IIIT Pets dataset across all evaluation settings. Although this dataset is small and prone to overfitting, Fourier-based augmentation effectively expands the data distribution, mitigating overfitting risks. Moreover, our method demonstrates noticeable improvement over the baseline on long-tailed fine-grained datasets such as Herbarium, maintaining competitive performance despite its class imbalance. 

\begin{wraptable}{r}{0.5\textwidth}
\vspace{-1em}
      \caption{\textbf{Comparison with state-of-the-art GCD methods on Herbarium19~\cite{tan2019herbarium} and Oxford-Pet~\cite{parkhi2012cats} on DINOv1.} Bold and underlined numbers indicate the best and second-best accuracies, respectively. Our method is well suited for fine-grained datasets, profits from stronger backbones, and has strong performance for all three experimental settings.}
  \centering
  \resizebox{1\linewidth}{!}{
\begin{tabular}{laccacc}
\toprule
&\multicolumn{3}{c}{\textbf{Oxford-Pet}}&\multicolumn{3}{c}{\textbf{Herbarium19}}\\ \cmidrule(lr){2-4} \cmidrule(lr){5-7} \textbf{Method}&All&Known&Novel&All&Known&Novel\\
\midrule
GCD \cite{vaze2022generalized}	&80.2 &85.1 &77.6& 35.4& 51.0& 27.0\\
SimGCD~\cite{wen2022simple}&91.7 &83.6& \underline{96.0}& 44.0& 58.0& 36.4\\
InfoSieve~\cite{rastegar2023learn} &91.8& \bf{92.6}& 91.3& 40.3& 59.0 &30.2\\
DebGCD~\cite{liu2025debgcd} &\bf{93.0}& 86.4 &\bf{96.5}& \underline{44.7}& \underline{59.4}& \underline{36.8}\\
SEAL~\cite{he2025seal} &\underline{92.9}& 88.9 &95.0 &\bf{46.9}& 45.8 &\bf{48.2}\\
\cmidrule{1-7}
SelEx~\cite{RastegarECCV2024} &92.5& \underline{91.9}& 92.8 &39.6& 54.9& 31.3\\
\rowcolor{gray!25}\textbf{FourEx (Ours)  } &92.8 &91.6& 93.4&44.5 &\bf{61.5}&35.3 \\
Avg $\Delta$&\textcolor{green!50!black}{+0.3}& \textcolor{red!60!black}{-0.3}&\textcolor{green!50!black}{+0.6}&\textcolor{green!50!black}{+4.9}&\textcolor{green!50!black}{+6.6}&\textcolor{green!50!black}{+4.0}\\
\bottomrule
\end{tabular}
}
  \label{tab:pets}
   \vspace{-1em}
\end{wraptable}

\begin{figure}[t]
  \centering
  \includegraphics[width=1\linewidth]{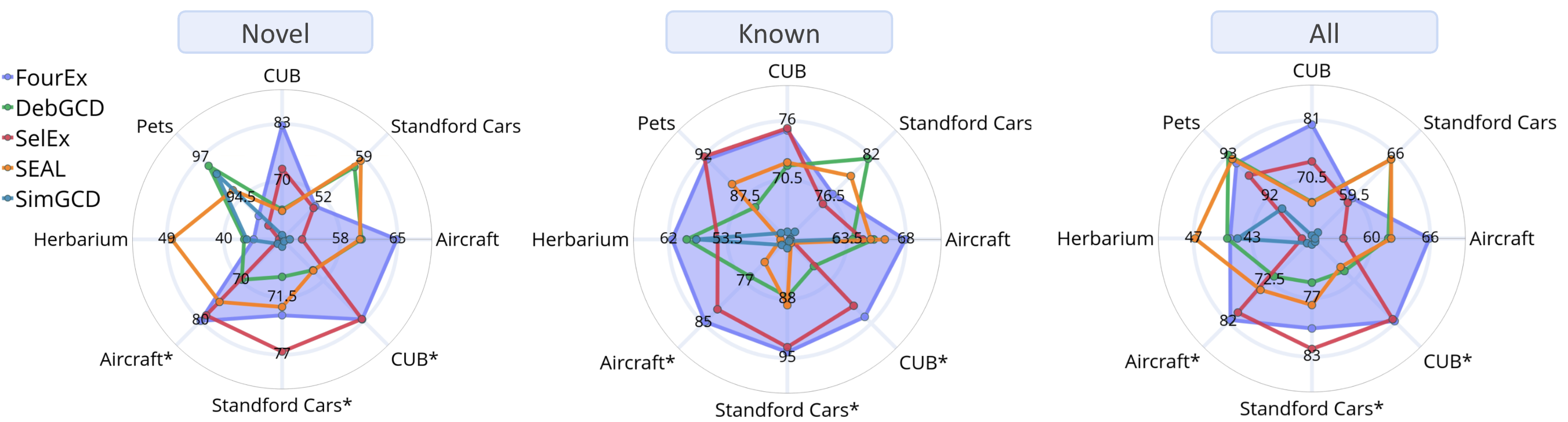}
   \caption{Radar chart comparing our proposed method against state-of-the-art approaches across five fine-grained datasets: CUB, Stanford Cars, Aircraft, Oxford Pets, and the long-tailed Herbarium. Datasets marked with $^*$ utilize the DINOv2 backbone. Our method visually envelops the baselines on Known and All metrics, while consistently outperforming the baseline, SelEx.}
   \label{fig:radar}
\end{figure}

\noindent\cref{fig:radar} illustrates the comparison between our method and state-of-the-art approaches across multiple datasets. As shown, while methodslike SEAL~\cite{he2025seal} achieve top performance on Stanford Cars and Herbarium, the radar chart reveals that these gains on novel categories come at the expense of accuracy on known categories. Similarly, although DebGCD~\cite{liu2025debgcd} leads on novel categories for Oxford Pets and known categories for Stanford Cars, our method consistently outperforms it in overall aggregate metrics. This demonstrates that our approach offers a more robust balance across diverse datasets with different granularities.


  \begin{wraptable}{r}{0.5\textwidth}
\vspace{-2.5em}
\caption{\textbf{Comparison with state-of-the-art with Estimated Number of Categories on DINOv1} Bold numbers show the best accuracies. We leverage the estimation technique from Vaze \emph{et al.} \cite{vaze2022generalized}
Our approach has competitive performance with other methods in all scenarios.}
  \centering
  \resizebox{1\linewidth}{!}{
\begin{tabular}{laccacc}
\toprule
&\multicolumn{3}{c}{\textbf{CUB-200}}&\multicolumn{3}{c}{\textbf{Stanford-Cars}}\\ \cmidrule(lr){2-4} \cmidrule(lr){5-7} \textbf{Method}&All&Known&Novel&All&Known&Novel\\
\midrule
GCD \cite{vaze2022generalized}	&47.1 &55.1 &44.8&39.1 &58.6& 29.7\\
SimGCD \cite{wen2022simple}&61.5& 66.4 &59.1&49.1& 65.1& 41.3\\
GCA~\cite{otholt2024guided}&62.0 &65.2& 60.4&54.4&72.1&45.8\\
$\mu$GCD \cite{vaze2023clevr4}&62.0 &60.3 &62.8& 56.3& 66.8& \bf{51.1}\\
PIM~\cite{tan2024interpret}&62.0& \underline{75.7} &55.1& 42.4 &65.3 &31.3\\
CMS \cite{choi2024contrastive}&64.4& 68.2& 62.2&51.7 &68.9 &43.4\\
SelEx \cite{RastegarECCV2024} &72.0 &72.3& 71.9& \underline{58.7}& \underline{75.3}& \underline{50.8}\\
Yang \textit{et. al.}~\cite{yang2024learning}&61.3& 60.8 &62.1 &44.3&58.2& 39.1\\
NC-GCD~\cite{hanconsistent}&\underline{70.3} &72.1& \underline{69.4}& 54.0& 73.1& 44.8\\ 
\cmidrule{1-7}
\rowcolor{gray!25}\textbf{FourEx (Ours)  } &\bf{77.3}& \bf{75.9} & \bf{77.9}& \bf{58.8}&\bf{77.6}&49.7 \\
\bottomrule
\end{tabular}
}

  \label{tab:est}
    \vspace{-3.5em}
\end{wraptable}

\noindent\textbf{Estimate the number of categories}. 
In our earlier experiments, we assumed prior knowledge of the exact number of categories, a condition that may not hold in practical applications. To evaluate the robustness of FourEx under more realistic scenarios, we tested its performance when the number of categories is estimated rather than known. We employed the estimation technique proposed by Vaze \textit{et al.} \cite{vaze2022generalized}, which yielded an estimated count of 231 categories for the CUB and 230 for Stanford-Cars dataset. 
As shown in \cref{tab:est}, FourEx outperforms existing methods across almost all evaluation metrics (\textit{All}, \textit{Known}, \textit{Novel}), demonstrating its effectiveness even when the category count is uncertain.

\noindent \textbf{Coarse-grained image classification.}
Finally, as detailed in the Appendix, while primarily designed for fine-grained scenarios, our approach also achieves competitive results on coarse-grained datasets, highlighting its versatility.

\subsection{Ablative studies}
In this section, we analyze the individual contributions of our method’s components and hyperparameters. All experiments are conducted on CUB-200 using the DINOv1 backbone. Additional ablations with other datasets, backbones, baselines, and qualitative results are provided in the Appendix.

\begin{table}[h!]
\centering
\caption{\textbf{Effectiveness of model components}. We evaluate the contribution of each model component across three fine-grained datasets. All modules contribute to improved performance on both known and novel categories. Note that the baseline is SelEx with the last three blocks fine-tuned.}

\resizebox{\columnwidth}{!}{%
\setlength{\tabcolsep}{6pt}{\small
\begin{tabular}{cccaccaccacca}
\toprule
\multicolumn{3}{c}{\textbf{Component}}&\multicolumn{3}{c}{\textbf{CUB-200}}&\multicolumn{3}{c}{\textbf{FGVC-Aircraft}}&\multicolumn{3}{c}{\textbf{Standford-Cars}}&\multicolumn{1}{c}{\textbf{Avg}}\\ 
\cmidrule(lr){1-3} \cmidrule(lr){4-6} \cmidrule(lr){7-9} 
\cmidrule(lr){10-12}\cmidrule(lr){13-13}
 $\mathbf{\mathcal{L}_\text{low}}$ &
   $\mathbf{\mathcal{L}_{\text{high}}}$ & $\mathbf{\mathcal{L}_{\text{cls}}}$   & \textbf{All}&\textbf{Known}&\textbf{Novel}&\textbf{All}&\textbf{Known}&\textbf{Novel}&\textbf{All}&\textbf{Known}&\textbf{Novel}&\textbf{All}\\
\midrule
& & &  76.4 &72.4 &78.4&61.2&67.7& 58.0& 56.9&76.9&47.3&64.8\\
 \gr{\ding{51}}  &  &  & 77.5& 74.0& 79.3&63.8& 67.7 &61.9&56.3& 77.1&46.3&65.9\\
  & \gr{\ding{51}} &  &78.8 &72.5 &81.9&63.9 &63.6&64.1&56.8 &74.0&48.5&66.5\\
  &  & \gr{\ding{51}} & 78.0& 71.2 &81.4&62.8 &65.5 &61.4& 57.1 &76.3 &47.8&66.0\\
 \gr{\ding{51}}  & \gr{\ding{51}} &  &80.0 &75.7&82.1&64.9 &69.6 &62.5
&57.7&78.2&47.8&67.5\\
 \gr{\ding{51}}  && \gr{\ding{51}} &77.9 &75.3 &79.2& 64.5& 66.3& 63.6
&57.7 &77.7 &48.0&66.7\\

 &\gr{\ding{51}}  & \gr{\ding{51}} &79.7 &72.8 &83.1&63.2 &66.1 &61.7
&57.9 &76.2 &49.1&66.9\\

\hline
\rowcolor{gray!25} \gr{\ding{51}}  & \gr{\ding{51}}&\gr{\ding{51}} & 80.2& 75.1& 82.7&65.9 &67.9&64.9&59.2 &76.9& 50.6&68.4\\
\bottomrule
\end{tabular}
}
}

\label{tab:multi}
\end{table}

\noindent\textbf{Effect of each component}.
\cref{tab:multi} examines the effect of our three key method components: low-frequency contrastive learning ($\mathcal{L}_{\text{low}}$), high-frequency contrastive learning ($\mathcal{L}_{\text{high}}$), and frequency augmentation classification ($\mathcal{L}_{\text{cls}}$).
The results demonstrate distinct benefits of low-frequency and high-frequency contrastive learning for novel and known categories, respectively. Our low-frequency contrastive learning, denoted as $\mathcal{L}_{\text{low}}$, shows a particular affinity for enhancing both known and novel categories. This aligns with our initial hypothesis that generalization for both set of categories benefit from more abstraction. Low-frequency signals create an implicit hierarchy in the training samples, necessitating correct categorization based on broad context. Conversely, our high-frequency contrastive learning, denoted as $\mathcal{L}_{\text{high}}$, excels in aiding novel categories. This is because hierarchical structures are advantageous for novel categories, but fine-grained details require the nuanced distinctions provided by high-frequency information. Interestingly, our classification objective alone tends to degrade performance for known categories. However, when decoupled with our contrastive losses, it improves results on novel categories. We attribute this to the prevention of overfitting, where the combined losses provide a more balanced training regimen.


\begin{table*}[ht]
  \caption{\textbf{Hyperparameter analysis.} These tables show the impact of each hyperparameter on the CUB dataset for both known and novel categories: (a) low-frequency coefficient $\alpha_{\text{low}}$ (b) high-frequency coefficient $\alpha_{\text{high}}$ (c) classification coefficient $\alpha_{\text{cls}}$ and (d) the SNR for the cutoff frequency.}
  \centering
  \resizebox{1\linewidth}{!}{

\begin{tabular}{cc}
    \begin{minipage}{.5\linewidth}
\resizebox{1\linewidth}{!}{
  \label{tab:comps}
\begin{tabular}{lacc}
\toprule

 \multicolumn{4}{c}{\textbf{(a) Effect of Low Frequency Hyperparameter}}\\ \cmidrule(lr){1-4}
$\textbf{Low Frequency Coef\ \ }$  &  All & Known  & Novel\\
\midrule
$\alpha_{\text{low}}=0$&77.9 &72.1&80.7\\
 $\alpha_{\text{low}}=0.01$ &80.2 &74.7 &83.0\\
 $\alpha_{\text{low}}=0.1$ & 80.2 &75.1 &82.7\\
 $\alpha_{\text{low}}=0.5$    & 76.8 &74.3 &78.0\\
 $\alpha_{\text{low}}=1$ &   69.2 &70.1 &68.8\\
\bottomrule
\end{tabular}}
  \end{minipage} &

\begin{minipage}{.5\linewidth}
\resizebox{0.95\linewidth}{!}{\begin{tabular}{lacc}
\toprule
 \multicolumn{4}{c}{\textbf{(b) Effect of High Frequency Hyperparameter}}\\ \cmidrule(lr){1-4}
\textbf{High Frequency Coef\ \ }  &  All & Known  & Novel\\
\midrule
 $\alpha_{\text{high}}=0$&78.1 &76.3 &78.9\\
 $\alpha_{\text{high}}=0.01$ &78.3 &75.8 &79.6
 \\
 $\alpha_{\text{high}}=0.1$ &  80.2 &75.1 &82.7 \\
 $\alpha_{\text{high}}=0.5$    & 79.2&75.0 &81.3\\
 $\alpha_{\text{high}}=1$ & 77.1& 73.7 &78.7
\\

\bottomrule
\end{tabular}}
  \end{minipage} \\
  &\\
\begin{minipage}{.5\linewidth}
\resizebox{0.95\linewidth}{!}{\begin{tabular}{lacc}
\toprule
 \multicolumn{4}{c}{\textbf{(c) Effect of Classification Hyperparameter}}\\ \cmidrule(lr){1-4}
\textbf{Classification Coef\qquad}  &  All & Known  & Novel\\
\midrule
 $\alpha_{\text{cls}}=0$  & 79.7 &76.5&81.4\\
 $\alpha_{\text{cls}}=0.01$ & 78.9& 73.1 &81.8  \\
 $\alpha_{\text{cls}}=0.1$    & 80.2 &75.1 &82.7 \\
$\alpha_{\text{cls}}=0.5$& 79.6 &73.3 &82.7\\
 $\alpha_{\text{cls}}=1$  & 73.1 &74.9 &72.3\\

\bottomrule
\end{tabular}}
  \end{minipage}&
  \begin{minipage}{.5\linewidth}
\resizebox{0.9\linewidth}{!}{\begin{tabular}{lacc}
\toprule
 \multicolumn{4}{c}{\textbf{(d) Effect of SNR of Cuttoff Frequency}}\\ \cmidrule(lr){1-4}
\textbf{SNR Threshold Choice\ }  &  All & Known  & Novel\\
\midrule
 SNR${=}$5 dB& 79.0 &75.7 &80.7\\
SNR${=}$10 dB&79.2 &72.3& 82.6\\
 SNR${=}$15 dB &80.2 &75.1 &82.7 \\
 SNR${=}$20 dB    & 78.4 &74.3 &80.5\\
 SNR${=}$25 dB&77.1 &74.3 &78.5\\

\bottomrule
\end{tabular}}
  \end{minipage}
  \\
\end{tabular}}

  \label{tab:allablations}

\end{table*}

\noindent\textbf{Effects of $\alpha_{\text{low}}$}. 
This hyperparameter represents the coefficient for the low-frequency component. As shown in \cref{tab:allablations}~(a) increasing $\alpha_{\text{low}}$ initially improves the model's performance on both novel and known categories. However, beyond a certain threshold, further increasing this hyperparameter begins to degrade performance. This behavior is intuitive, as the model relies on a blurred version of the data for this loss; excessive emphasis on low-frequency information causes the model to lose critical fine-grained details necessary for accurate categorization. As expected, the positive impact of this hyperparameter is more pronounced on novel categories, where the generalization facilitated by low-frequency components plays a larger role.

\noindent\textbf{Effects of $\alpha_{\text{high}}$}. 
This hyperparameter represents the coefficient for high-frequency components. As shown in \cref{tab:allablations}~(b), increasing $\alpha_{\text{high}}$ improves the model's performance on novel categories, while this hyperparameter negatively impacts known categories performance. This behavior aligns with the fact that high frequencies inherently carry lower energy compared to the overall image. When the model overemphasizes these low-energy components, it struggles to maintain stability when integrating the higher-energy low-frequency components, leading to degraded performance.

\noindent\textbf{Effects of $\alpha_{\text{cls}}$}. 
In our experiments, $\alpha_{\text{cls}}$ represents the classification coefficient applied to both low-frequency and high-frequency constructed images. As shown in \cref{tab:allablations}~(c), increasing this hyperparameter improves performance for novel categories. However, this trend diminishes beyond a certain threshold for $\alpha_{\text{cls}}$, indicating that the model might overfit to the low/high-frequency reconstructions. 


\begin{figure*}[t!]
  \centering
  \includegraphics[width=\linewidth]{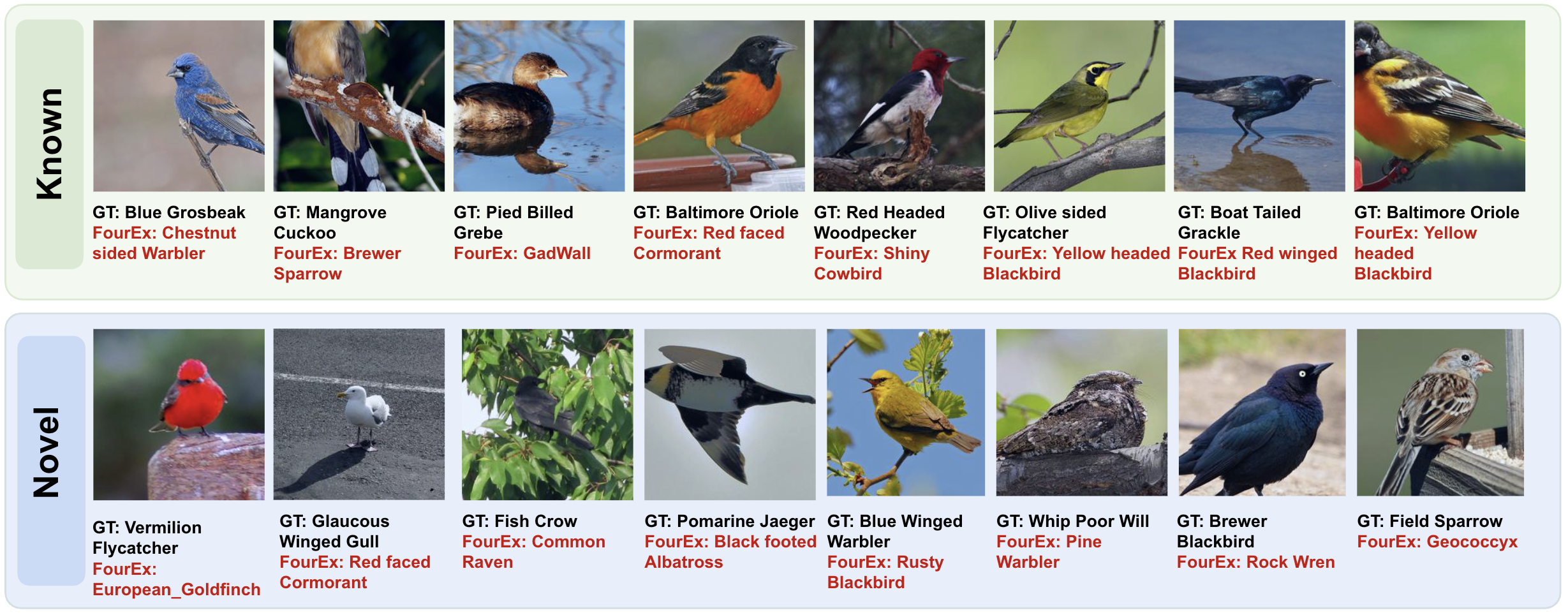}
 \vspace{-0.5em}
\caption{\textbf{Failure Cases of Our Model.} Many errors arise from occlusion, reflection, or color mismatches. While the model misclassifies in these examples, it often identifies a related species within the same hierarchy.} 
\label{fig:fails}
\end{figure*}

\noindent\textbf{Effects of Different SNRs}. 
In our experiments, we primarily adopt an SNR of 15 dB, as it represents the standard minimum threshold for acceptable subjective quality in most practical scenarios. To further assess the impact of signal quality on model performance, we evaluate the model across a range of SNR values. The results, summarized in \cref{tab:allablations}~(d), demonstrate the model's adaptability to varying SNR levels. Notably, lower SNRs tend to favor generalization to novel categories, while higher SNRs improve adaptation to known categories. However, we observe that beyond 20 dB, the performance gains plateau, indicating that increasing resolution beyond this threshold does not significantly enhance the model's ability to distinguish categories. This suggests a practical limit to the benefits of resolution in generalized category discovery.

\noindent\textbf{Failure Cases}.
\Cref{fig:fails}, illustrates instances where our model misclassified the data. A notable number of these errors are attributed to reflection or occlusion. Additionally, color appears to be a contributing factor. Since our model's high-pass filter removes color information, it becomes more susceptible to these types of misclassifications.

\section{Conclusion}
In this paper, we introduce a frequency-based augmentation strategy for self-supervised learning aimed at improving fine-grained generalized category discovery. By leveraging the complementary roles of low- and high-frequency information, our approach enables models to capture both broad, abstract patterns that support generalization and subtle details essential for precise discrimination. Low-frequency contrastive learning encourages the emergence of implicit category hierarchies, while high-frequency learning enhances sensitivity to fine-grained distinctions. Our method consistently improves performance on fine-grained benchmarks and achieves competitive results on long-tailed and coarse-grained datasets (see Appendix), demonstrating robustness across varying levels of category granularity. Moreover, the approach is plug-and-play and can be readily integrated into existing architectures with minimal additional computational overhead, making it practical and widely applicable.

\section*{Acknowledgments}
{This work is part of the project Real-Time Video Surveillance Search with project number 18038, which is (partly) financed by the Dutch Research Council (NWO) domain Applied and Engineering/Sciences (TTW).}

{
    \small
    \bibliographystyle{splncs04}
    \bibliography{main}
}

\section{Motivation}
\begin{figure}[t!]
\includegraphics[width=\linewidth]{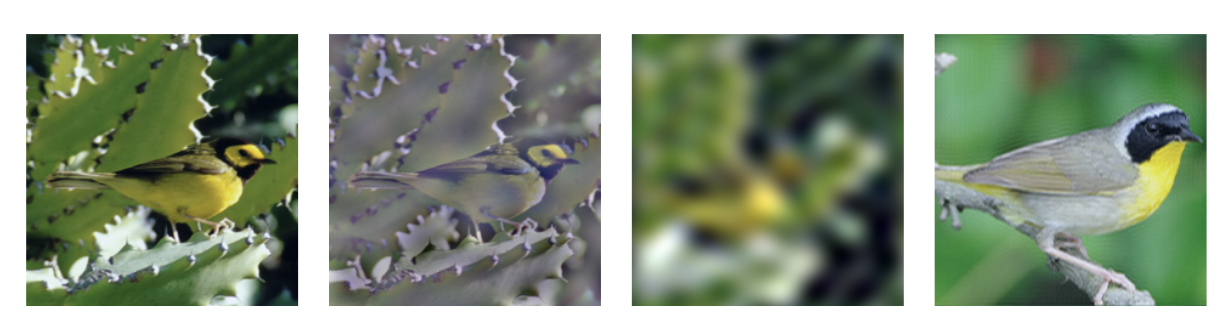}
\vspace{-1.5em}
\caption{{\textbf{Which one is the different one?} The first from the left image is the original photo. The second image shows a high-frequency reconstruction of the original. The third image represents a low-frequency reconstruction of the original photograph. In contrast, the right image, despite its visual similarity to the original, is distinct from the other images presented. }
}\label{fig:fig1}
\end{figure}

\begin{figure*}[t!]
\vspace{0em}
  \centering
  \includegraphics[width=\linewidth]{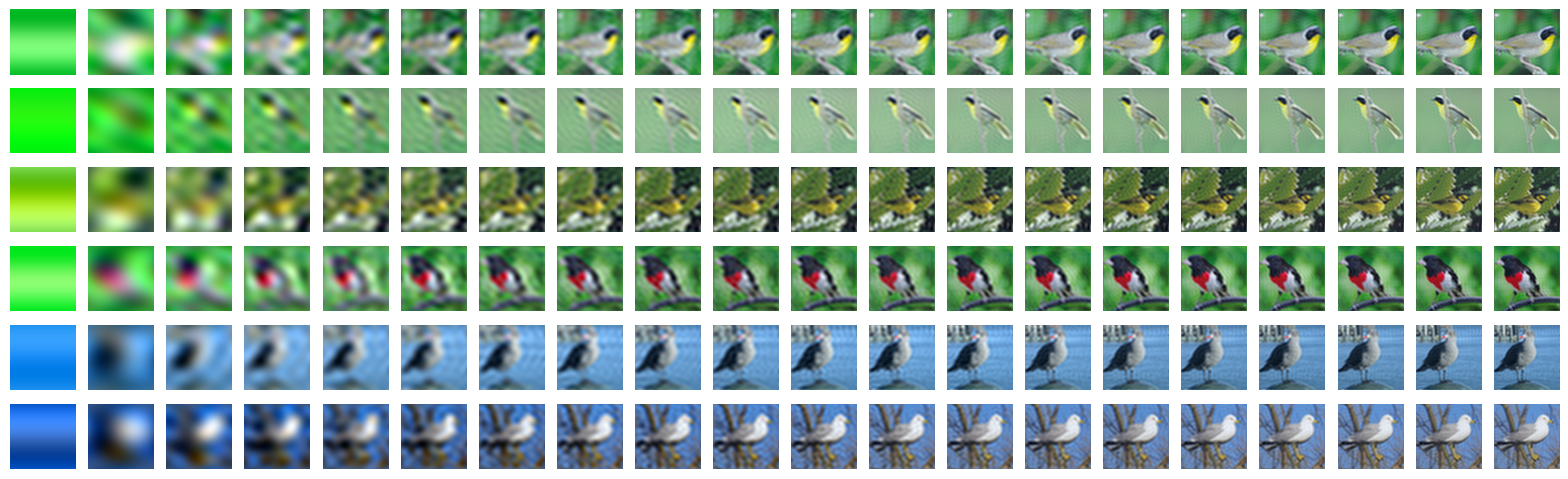}
\caption{\textbf{Motivation for Fourier Self-Supervision with Low Frequencies.} From left to right, we gradually increase the cut-off frequency, retaining more low-frequency components. These components capture broad category traits, allowing blurred images to hint at the environmental context (e.g., marine vs. forest birds) even without precise details. This hierarchy enables early generalization over seen categories, supporting novel class discrimination. By progressively compressing low-frequency information into initial latent dimensions, the model reserves later dimensions for fine-grained distinctions. } 
\label{fig:fourlow}

\end{figure*}

\begin{figure*}[t!]
\vspace{0em}
  \centering
  \includegraphics[width=\linewidth]{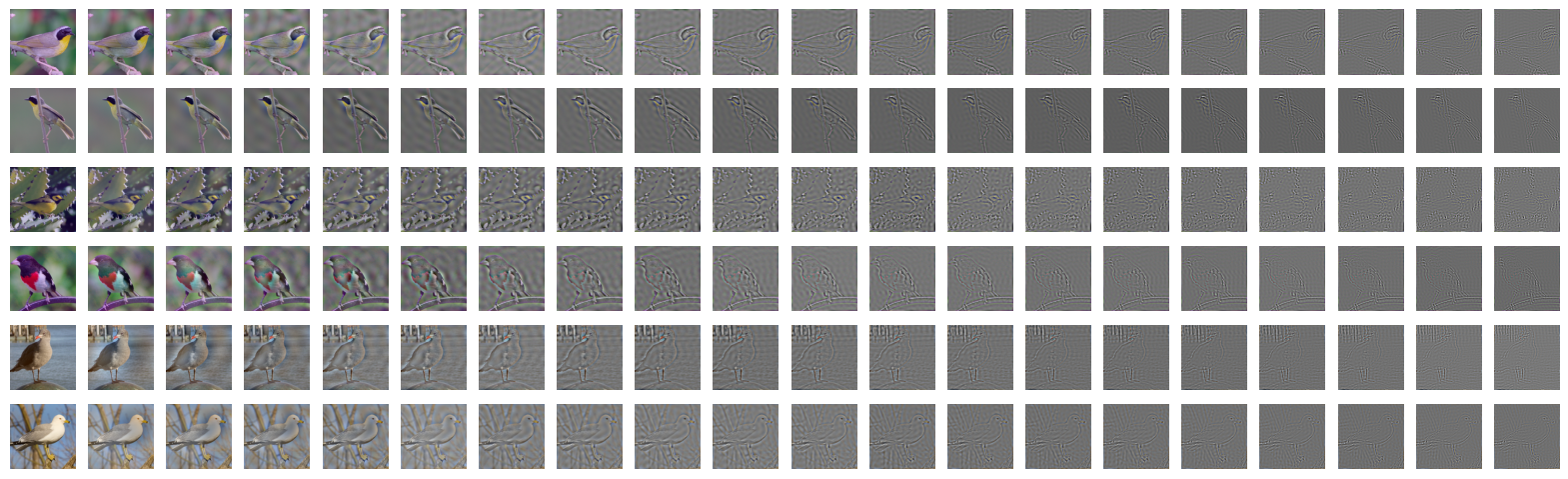}
\caption{\textbf{Motivation for High-Frequency Fourier Self-Supervision.} From left to right, we progressively reduce low-frequency components, retaining only high frequencies. High-frequency components capture fine-grained distinctions, such as the shape of a bird's beak or eye, which are less influenced by broad environmental cues. By dedicating parts of the latent dimension to higher frequencies, we encourage the model to base decisions on detailed shape information rather than texture, enhancing its capacity for fine-grained categorization. Since high frequencies are less dependent on environmental context, this approach also strengthens the model's ability to generalize to novel categories. }
\label{fig:fourhigh}
\end{figure*}

To address the challenge of fine-grained generalized category discovery (GCD), we draw inspiration from Fourier analysis. Intuitively, object categorization should remain consistent across different frequency bands of its Fourier transform. Lower frequencies often represent more generalized, abstract features, making them suitable for identifying broad categories, while higher frequencies capture finer details that are crucial for distinguishing visually similar classes. This perspective guides our exploration of frequency-based self-supervision as a means to enhance concept discovery, particularly in scenarios where classes exhibit high visual similarity. Unlike machine learning models, human cognition does not require extensive exposure to varied examples of an object to recognize it under different conditions. Consider the bird images in \cref{fig:fig1}. While the left and right images share visual similarities, human perception effortlessly identifies the right one as distinct from the other three, which represents the same entity altered by Fourier frequency filtering. This natural capability stems from how the human visual system processes light frequencies, where the eye's lens effectively performs a Fourier transform on the incoming light, forming an image on the retina~\cite{goodman2005introduction}. Inspired by this physiological phenomenon, we hypothesize that consistent categorization across different frequency bands of an object's Fourier transform could enhance both generalization and discrimination. Lower frequencies support more abstract category recognition as shown in~\cref{fig:fourlow}, while higher frequencies aid in distinguishing fine-grained differences as depicted in~\cref{fig:fourhigh}.

\section{Approach Specification}
In this section, we demonstrate how to integrate our Fourier-based method with SelEx \cite{RastegarECCV2024} and GCD \cite{vaze2022generalized} in more details. For integration with GCD, we can simply replace $\mathcal{L}_{\text{base}}$ with $\mathcal{L}_{\text{GCD}}$ as presented in the main text. For SelEx, there are a few nuances that we detail in this section. Consider an image $\mathbf{x}_i$, its low-frequency reconstruction which is denoted by $\mathbf{l}_i$ is as follows:
\begin{equation}
    \label{eq:lowrec}
    \mathbf{l}_i{=}\mathcal{F}^{-1}(H_{LP}\odot\mathcal{F}(\mathbf{x}_i)).
\end{equation}
As in the GCD integration, for SelEx, we feed the reconstructed image into both its supervised and unsupervised self-expertise modules. However, since both SelEx and our method restrict feature dimensions, we take the minimum between the Fourier reconstruction fraction of the dimension and its pseudo-label representation counterpart. The pseudo-labels generated by the SelEx hierarchical extraction process serve as supervisory signals in each epoch. Denote by $\mathcal{L}s^k$ the supervised contrastive loss at pseudo-label level $k$.  The total supervised contrastive loss in SelEx is:
\begin{equation}
    \label{eq:sup}
\mathcal{L}_{\text{SSE}}={1\over2}(\sum_{k=0}^{\lg K}\frac{\mathcal{L}_s^k|{D\over 2^k}}{2^k}),
\end{equation}
For incorporating Fourier loss, we are considering $\frac{f_{l}}{T}$ of the latent dimension $D$. We define this fraction as $d {=} \frac{f_{l}}{T}D$ and incorporate it into \cref{eq:sup} to compute the low-frequency contrastive loss:
\begin{equation}
    \label{eq:sup}
\mathcal{L}_{\text{LSSE}}={1\over2}(\sum_{k=0}^{\lg K}\frac{\mathcal{L}_s^k|\min({D\over 2^k},d)}{2^k}),
\end{equation}
The high-frequency component uses the remaining portion of the representation.     
\section{Experiments}

\subsection{Experimental Setup}
{
\begin{table}[t]
\caption{\textbf{Statistics of datasets and their data splits for the generalized category discovery.} The first three datasets are coarse-grained, and the next four datasets are fine-grained. The Herbarium19 dataset is both fine-grained and long-tailed.}
\vspace{-1em}
\begin{center}
\begin{sc}
\resizebox{0.9\linewidth}{!}{
\begin{tabular}{lrrrr}
\toprule
& \multicolumn{2}{c}{\textbf{Labelled}} & \multicolumn{2}{c}{\textbf{Unlabelled}}\\ \cmidrule(lr){2-3} \cmidrule(lr){4-5}
\textbf{Dataset}  & \#Images & \#Categories   &  \#Images & \#Categories  \\
\midrule
CIFAR-10 \cite{krizhevsky2009learning}       & 12.5K & 5 & 37.5K & 10   \\  
CIFAR-100 \cite{krizhevsky2009learning}       & 20.0K & 80 & 30.0K & 100 \\ 
ImageNet-100 \cite{deng2009imagenet}& 31.9K& 50& 95.3K& 100\\
\midrule
CUB-200 \cite{wah_branson_welinder_perona_belongie_2011} & 1.5K & 100 & 4.5K &200 \\
Stanford-Cars \cite{krause20133d}& 2.0K & 98 & 6.1K &196 \\ 
FGVC-Aircraft \cite{maji2013fine}& 3.4K& 50& 6.6K& 100\\
Oxford-Pet \cite{parkhi2012cats}& 0.9K& 19& 2.7K& 37\\
\midrule
Herbarium19 \cite{tan2019herbarium}& 8.9K& 341& 25.4K& 683\\

\bottomrule
\end{tabular}}
\end{sc}
\end{center}

\label{tab:data}
\end{table}
}

\textbf{Datasets.}
We evaluate the performance of our proposed method across several datasets to demonstrate its effectiveness and versatility. We conduct experiments on four fine-grained datasets: Caltech-UCSD Birds-200-2011 (CUB-200)~\cite{wah_branson_welinder_perona_belongie_2011}, Fine-Grained Visual Classification of Aircraft (FGVC-Aircraft)~\cite{maji2013fine}, Stanford Cars~\cite{krause20133d}, and Oxford-IIIT Pet~\cite{parkhi2012cats}. These datasets assess the method's efficacy in fine-grained image classification tasks, which involve identifying subtle distinctions among highly similar categories, such as different bird species, aircraft models, car brands, and pet breeds, thus posing significant challenges. Additionally, we extend our evaluation to more general coarse-grained datasets, including CIFAR10, CIFAR100~\cite{krizhevsky2009learning}, and a subset of ImageNet comprising 100 categories, referred to as ImageNet-100~\cite{deng2009imagenet}. This extension highlights the adaptability of our method to broader classification challenges beyond fine-grained tasks. CIFAR10 and CIFAR100 feature coarse-grained categories encompassing a wide range of objects, such as vehicles and animals, while ImageNet-100 provides a diverse set of general categories from the larger ImageNet dataset. Finally, we use the challenging Herbarium dataset by Tan~\etal~\cite{tan2019herbarium}, which is fine-grained and long-tailed, to show that this approach can still be beneficial even if the dataset is unbalanced.
To provide a comprehensive overview of our experiments, we include a detailed summary of the datasets, including their statistical distribution and train/test splits, in \cref{tab:data}.

\noindent\textbf{CIFAR10/100}~\cite{krizhevsky2009learning} are generic datasets encompassing a variety of general object categories, including vehicles and animals, highlighting the diverse challenges in coarse-grained image classification.

\noindent\textbf{ImageNet-100}~\cite{deng2009imagenet}  is a representative subset of the extensive ImageNet database, focusing on 100 varied categories to evaluate the generalizability of our approach in a broader context.

\noindent\textbf{CUB-200}~\cite{wah_branson_welinder_perona_belongie_2011}, also known as Caltech-UCSD Birds-200-2011, emphasizes the importance of discerning minute details across different bird species, underscoring the complexity of fine-grained image recognition tasks.

\noindent\textbf{FGVC-Aircraft}~\cite{maji2013fine} focuses on aircraft, where variations in design significantly affect structural appearances, highlighting the dataset's fine-grained nature.

\noindent\textbf{Stanford Cars}~\cite{krause20133d} introduces the challenge of recognizing car brands from various angles and in different colors.

\noindent\textbf{Oxford-IIIT Pet}~\cite{parkhi2012cats} centers on the classification of cat and dog breeds, where limited data availability increases the risk of overfitting, stressing the dataset's specific challenge.

\noindent\textbf{Herbarium}~\cite{tan2019herbarium}, focuses on plant species and features a long-tailed distribution, illustrating the application of fine-grained classification when categories are unbalanced.

\noindent\textbf{Implementation Details.}
In our experiments, we followed the dataset division proposed by Vaze~\etal~\cite{vaze2022generalized}, where half of the categories in each dataset are designated as known. The labeled set consists of 50\% of the samples from these known categories. The remainder of the known category data, along with all data from novel categories, comprise the unlabeled set.

Following~\cite{vaze2022generalized}, we use ViT-B/16 as our backbone, pre-trained either by DINOv1~\cite{caron2021emerging} on unlabeled ImageNet-1K~\cite{krizhevsky2017imagenet}, or by DINOv2~\cite{oquab2024dinov} pre-trained on the LVD-142M dataset. We use a batch size of 128 for training. Unlike Vaze~\etal~\cite{vaze2022generalized}, we froze the first 9 blocks of ViT-B/16 for FourEx and fine-tuned the last three blocks, to leverage more parameters given that for each component, only a fraction of the latent dimension is considered. For FourSim and FourGCD, we stick to their default of finetuning only the last block. We refer to our SelEx-based variant as FourEx, as discussed in the main paper. For the version integrated into the GCD pipeline, which we present in the appendix, we use the name FourGCD.

Before training, we measure the signal-to-noise ratio (SNR) across a range of cutoff frequencies and select the cutoff that achieves an SNR of 15 dB. The corresponding cutoff values for different datasets and SNR levels are provided in the \cref{tab:SNRdiff}. To further mitigate Gibbs ringing artifacts that occur near edges, we apply a Gaussian blur with a kernel size of 5 to each image before filtering.
\begin{table}[ht]

 \caption{\textbf{Calculated Cutoff Frequency for different SNRs.} For each dataset, we determine the maximum cutoff frequency that results in a particular SNR.
 }
  \centering
\resizebox{0.7\linewidth}{!}{
\begin{tabular}{lccccc}

\toprule
\multicolumn{5}{c}{\textbf{Cutoff Frequency for Different SNRs}}\\ \cmidrule(lr){1-6}
 SNR&
   5dB &  10dB &  15dB & 20dB & 25dB\\
\midrule
CUB-200 \cite{wah_branson_welinder_perona_belongie_2011} & 2& 8& 23& 45&55 \\
FGVC-Aircraft \cite{maji2013fine}& 2 & 9 & 27 &50&55\\
Stanford-Cars \cite{krause20133d}& 3 & 11 & 27 &47&55 \\ 
Oxford-Pet \cite{parkhi2012cats}& 2& 7&18&38&55\\
Herbarium19 \cite{tan2019herbarium}& 1&2 & 13& 32&50\\
         
\bottomrule
\end{tabular}}

   \label{tab:SNRdiff}
   \vspace{0em}
\end{table}

\noindent\textbf{Evaluation Metrics.}
We employ Semi-Supervised $k$-means to cluster the extracted embeddings using a $k$-means approach. To assign the emerging clusters to their ground truth labels optimally, we utilize the Hungarian algorithm \cite{wright1990speeding}. We then report the model's prediction accuracy across three categories: \textit{All}, \textit{Known}, and \textit{Novel}. Accuracy on \textit{All} is calculated using the entire unlabeled test set, which includes both known and unknown categories. For \textit{Known}, we consider samples with labels that were known during training. For \textit{Novel}, we evaluate samples from the unlabeled categories present at train time.

\begin{table*}[t!]
      \caption{\textbf{Comparison with improvement over different backbones for fine-grained image classification.} Our approach improves on its baseline for all three experimental settings (\textit{All}, \textit{Known}, and \textit{Novel}).}
  \centering

  \resizebox{1\linewidth}{!}{
\begin{tabular}{claccaccaccacc}
\toprule
&&\multicolumn{3}{c}{\textbf{CUB-200}}& \multicolumn{3}{c}{\textbf{FGVC-Aircraft}}&\multicolumn{3}{c}{\textbf{Stanford-Cars}}&\multicolumn{3}{c}{\textbf{Oxford-IIIT Pet}}\\ \cmidrule(lr){3-5} \cmidrule(lr){6-8} \cmidrule(lr){9-11} \cmidrule(lr){12-14}&\textbf{Method}&All&Known&Novel&All&Known&Novel&All&Known&Novel&All& Known&Novel\\
\bottomrule
\rowcolor{gray!25}&\multicolumn{3}{l}{\textbf{DINOv1 Backbone}}&&&&&&&&&&\\
&GCD \cite{vaze2022generalized}	&51.3&	56.6&	48.7&	45.0	&41.1&46.9&39.0&57.6&29.9&80.2	&85.1&77.6\\
&\textbf{FourGCD}            &69.4& 75.8 &66.2&59.4 &67.6 &55.3&57.6 &79.7&46.9&91.9 &91.9 &91.8\\
\cmidrule{2-14}
&SelEx~\cite{RastegarECCV2024} &73.6 &75.3 &72.8&57.1 &64.7&53.3& 58.5 &75.6&50.3&92.5 & 91.9&92.8\\
&\textbf{FourEx  } &80.2& 75.1& 82.7&65.9 &67.9&64.9&59.2 &76.9& 50.6&92.8 &91.6& 93.4 \\

\bottomrule
\rowcolor{gray!25}&\multicolumn{3}{l}{\textbf{DINOv2 Backbone}}&&&&&&&&&&\\
&GCD \cite{vaze2022generalized}&71.9& 71.2& 72.3&55.4 &47.9& 59.2& 65.7&67.8&64.7&87.6 &88.2& 87.2\\
&\textbf{FourGCD}            &76.4 &81.7 &73.8& 77.1 &81.1 &75.1&  78.2&91.6 & 71.8&94.1 &95.7 &93.3\\
\cmidrule{2-14}

&SelEx \cite{RastegarECCV2024}            &87.4&85.1&88.5&79.8&82.3 &78.6&  82.2 &93.7&76.7&95.1 &95.9 &94.8\\
&\textbf{FourEx}   & 87.8 &86.3 &88.6
 &  81.5 &84.7 &80.0
&  80.1 &94.3 &73.2 &95.2 &95.6 &95.0\\
\bottomrule
\end{tabular}
}

  \label{tab:cubsupp}
\end{table*}

\subsection{Comparison with State-of-the-Art}
\noindent \textbf{Fine-grained image classification.}
In \cref{tab:cubsupp}, we evaluate our two variants (FourEx, and FourGCD) improvement over their corresponding baselines (SelEx, GCD) on four fine-grained datasets. Our method demonstrates strong performance, improving on both baselines in three \textit{All} and \textit{Novel} and \textit{known} category classifications. This advantage primarily stems from the model's capability to distinguish subtle high-frequency details from more general low-frequency features, critical for accurately classifying closely related categories requiring nuanced recognition. Furthermore, by incorporating FourEx, which leverages SelEx as its base model, we observe enhanced synergy between frequency-domain information and hierarchical granularity, leading to further performance improvements. 
\\

\noindent \textbf{Novel category discovery efficacy.}  
Our performance gains on novel categories vary across datasets due to how discriminative information is distributed across frequencies. However, FourEx consistently improves over SelEx on novel categories, with +5.3\% on DINOv1, -0.7\% on DINOv2, and +3.1\% overall. Datasets relying on subtle details (CUB, Aircraft) benefit more, while coarse-shape datasets (Stanford Cars) show smaller but positive gains. Notably, on DINOv1, FourEx leads over the second-best method by +4.0\% on three fine-grained datasets.
\\

\begin{table}[h!]
\caption{FourEx improvement on novel category discovery. We can see that the more a dataset is fine-grained, the more improvement our approach makes. 
}\centering
\resizebox{\columnwidth}{!}{%
\setlength{\tabcolsep}{7pt}{\small
\begin{tabular}{lcccccccccc}
\toprule

\textbf{Dataset}  &\textbf{CUB}&\textbf{Aircraft}&\textbf{Scars}&\textbf{CUB$^*$}&\textbf{Aircraft$^*$}&\textbf{Scars$^*$}&\textbf{Pets}&\textbf{Herb}&\textbf{All}\\
\midrule

SelEx (Novel) &72.8&53.3&50.3&88.5&78.6&76.7&92.8&31.3&68.0\\
FourEx (Novel) &82.7&64.9&50.6&88.6&80.0 &73.2&93.4& 35.3&71.1 \\
\hline
\rowcolor{gray!25}$\Delta$&+9.9&+11.6&+0.3&+0.1&+1.4 &-3.5&+0.6 &+4.0&+3.1\\

\bottomrule
\end{tabular}
}
}

\label{tab:simgcd}
\end{table}
\noindent\textbf{Coarse-grained image classification.}
We also evaluate both FourEx and FourGCD on three coarse-grained datasets: CIFAR10/100 ~\cite{krizhevsky2009learning} and ImageNet-100~\cite{deng2009imagenet}. \cref{tab:cifar10supp}  provides a comparative analysis of our method against existing state-of-the-art generalized category discovery methods. While developed for fine-grained category identification, our method also exhibits competitive results on coarse-grained datasets for both \textit{Known} and \textit{Novel} categories. Despite the reduced need for fine-grained distinctions in these datasets, FourGCD and FourEx outperform baseline methods such as GCD~\cite{vaze2022generalized} and SelEx~\cite{RastegarECCV2024}, with FourEx showing particularly competitive overall performance

\begin{table*}[ht!]
  \caption{\textbf{Comparison with state-of-the-art for coarse-grained image classification.} Bold and underlined numbers show the best and second-best accuracies. Our method has a consistent performance for the three experimental settings (\textit{All}, \textit{Known}, \textit{Novel}). Our method is especially suitable for known categories in all three datasets. }
  \label{tab:cifar10supp}
  \centering
  \resizebox{1\linewidth}{!}{
\begin{tabular}{claccaccaccacc}
\toprule
&&\multicolumn{3}{c}{\textbf{CIFAR-10}}  & \multicolumn{3}{c}{\textbf{CIFAR-100}}&\multicolumn{3}{c}{\textbf{ImageNet-100}}&\multicolumn{3}{c}{\textbf{Average}}\\ \cmidrule(lr){3-5} \cmidrule(lr){6-8} \cmidrule(lr){9-11} \cmidrule(lr){12-14}
&\textbf{Method} &All & Known  & Novel &  All & Known  & Novel&All & Known  & Novel&All & Known  & Novel\\
\midrule
\multirow{10}{*}{\rotatebox{90}{DINOv1}}
&GCD \cite{vaze2022generalized}    & 91.5 & 97.9& 88.2 & 73.0 & 76.2  & 66.5&74.1 & 89.8&	66.3&79.5 & 88.0  &73.7 \\

&GPC \cite{zhao2023learning} &90.6& 97.6& 87.0& 75.4 &84.6& 60.1&75.3& 93.4 & 66.7&80.4 &\underline{91.9}& 71.3\\
&XCon \cite{fei2022xcon} & 96.0&97.3&95.4&74.2&81.2&60.3&77.6& 93.5 &69.7&82.6&90.7&75.1\\
&ORCA$^\dag$ \cite{cao2021open}         & 96.9 & 95.1 & 97.8 & 74.2 & 82.1  & 67.2&79.2	&93.2	&72.1&83.4&90.1&79.0\\
&DCCL \cite{pu2023dynamic}        & 96.3 & 96.5 & 96.9 & 75.3 & 76.8  & 70.2&80.5&90.5	&76.2&84.0 & 87.9  & 81.1\\
&InfoSieve \cite{rastegar2023learn}                  &94.8&97.7 &93.4
& 78.3& 82.2 &70.5& 80.5 &93.8 &73.8&84.5& 91.2 &79.2\\
&SimGCD \cite{wen2022simple}       & 97.1 & 95.1 &98.1 &80.1 & 81.2 & \underline{77.8}& 83.0 &93.1&77.9&86.7 & 89.8& 84.6\\

&PromptCAL \cite{zhang2022promptcal} &\bf{97.9}& 96.6& \underline{98.5}& 81.2 & 84.2 & 75.3&83.1	&92.7	&78.3 &87.4 & 91.2 & 84.0\\
&SPTNet \cite{wang2024sptnet} &\underline{97.3}&95.0&\bf{98.6}& 81.3&\underline{84.3}&75.6&\underline{85.4}&93.2&81.4&\bf{88.0}&90.8&\bf{85.2}\\
&SelEx \cite{RastegarECCV2024}& 95.9&\bf{98.1} &94.8&\bf{82.3}&\bf{85.3}&76.3&83.1& 93.6 &77.8&87.1& \bf{92.3}&83.0\\
\cmidrule{1-14}

\rowcolor{gray!25}&\textbf{FourGCD (Ours)}         &95.1&\underline{98.0} &93.6&80.3 &83.3 &74.3
&81.2&\underline{94.3} &74.6&85.5& 91.8&80.8\\
\rowcolor{gray!25}&\textbf{FourEx (Ours)}   &96.8 &97.2 &96.7&\underline{81.5} &83.3 &\bf{77.9}& \bf{86.1} &\bf{94.7} &81.8&\underline{87.7}&91.5&\underline{85.0} \\
\bottomrule
\end{tabular}
}
\end{table*}

\subsection{Experimental Analysis}
\noindent \textbf{Time-Complexity Analysis.} 
We analyze the time complexity of our two variants, which is influenced by the Fourier and inverse Fourier transforms performed during each epoch. Utilizing GPU-based FFT, where the entire image is loaded into memory, the computational complexity is $\mathcal{O}(\log N)$, where $N$ represents the image size. To provide a practical perspective, the timing results for the GPU implementation are reported in \cref{tab:time}.
\begin{table}[h!]
\caption{Time and Computational complexity of FourGCD and FourEx in comparison with their corresponding baselines GCD and SelEx. Our approach introduces a low overhead but strong improvements over the baselines. }
\vspace{0em}
\centering
{\resizebox{0.8\linewidth}{!}{%
\begin{tabular}{lcccccc}
\toprule
&\multicolumn{3}{c}{\textbf{Accuracy on CUB}}&\multicolumn{3}{c}{\textbf{Estimated Performance}}\\ 
\cmidrule(lr){2-4} \cmidrule(lr){5-7} 
\textbf{Method}&All&Known&Novel &Epoch time & FLOPs & Params \\
\midrule
GCD (Baseline)&51.3&56.6&48.7   & 61.57 s &19.98 G &37.25 M\\
FourGCD &69.4&75.8&66.2  &82.27 s  &19.98 G &37.25 M\\
SelEx (Baseline) &73.6	&75.3&	72.8  &84.29 s  &40.43 G&37.25 M\\
FourEx &80.2 &75.1 &82.7&90.69 s  &40.43 G&37.25 M\\
\bottomrule
\end{tabular}
}
}
\vspace{0em}

\label{tab:time}
\end{table}
\\

\noindent\textbf{Estimate the number of categories}. 
In our earlier experiments, we assumed prior knowledge of the exact number of categories, a condition that may not hold in practical applications and we showed that FourEx can achieve SoTA results even when the number of categories are uncertain. To evaluate the robustness of other variant FourGCD under unknown category number scenario, we tested its performance when the number of categories is estimated rather than known. We employed the estimation technique proposed by Vaze \textit{et al.} \cite{vaze2022generalized}, which yielded an estimated count of 231 categories for the CUB and 230 for Stanford-Cars dataset. It is important to note that the category counts estimated by methods like DCCL \cite{pu2023dynamic} and GCA \cite{otholt2024guided} differ from those obtained using other techniques. 
As shown in \cref{tab:estsupp}, FourGCD improves on the baseline and has competitive performance against existing methods across almost all evaluation metrics (\textit{All}, \textit{Known}, \textit{Novel}), demonstrating its effectiveness even when the category count is uncertain.

\begin{table}[ht!]
\caption{\textbf{Comparison with Estimated Number of Categories} Bold numbers show the best accuracies. We leverage the estimation technique from Vaze \emph{et al.} \cite{vaze2022generalized} to determine a category count of 231 for CUB. Note that the category estimations by DCCL \cite{pu2023dynamic} and GCA \cite{otholt2024guided} diverge from other methods. Our approach has competitive performance with other methods in all scenarios (\textit{All}, \textit{Known}, \textit{Novel}), showing robustness even with estimated category numbers.}  
  \centering
  {\resizebox{0.6\linewidth}{!}{
\begin{tabular}{clcacc}
\toprule
&&& \multicolumn{3}{c}{\textbf{CUB-200}}\\ \cmidrule(lr){4-6} 
&\textbf{Method} &\#Classes & All & Known  & Novel \\
\midrule
\multirow{6}{*}{\rotatebox{90}{DINOv1}}&GCD \cite{vaze2022generalized} &Known& 51.3&	56.6&	48.7\\

&DCCL \cite{pu2023dynamic}        &Known&63.5&60.8&64.9\\

&SimGCD \cite{wen2022simple}       &Known&60.3&65.6&57.7\\

&GCA~\cite{otholt2024guided}&Known&\underline{68.8}&\underline{73.4}&\bf{66.6}\\

&$\mu$GCD~\cite{vaze2023clevr4}&Known&65.7&68.0&64.6\\
\midrule
\rowcolor{gray!25}&\textbf{FourGCD (Ours)} &Known&\bf{69.4}& \bf{75.8} &\underline{66.2}\\
\midrule
\multirow{6}{*}{\rotatebox{90}{DINOv1}}&GCD \cite{vaze2022generalized} &Estimated& 47.1&55.1&44.8\\

&DCCL \cite{pu2023dynamic}        &Estimated&\underline{63.5}&60.8&\underline{64.9}\\

&SimGCD \cite{wen2022simple}  & Estimated&61.5&\underline{66.4}&59.1\\

&GCA~\cite{otholt2024guided}&Estimated&62.0&65.2&60.4\\

&$\mu$GCD~\cite{vaze2023clevr4}&Estimated&62.0&60.3&62.8\\
\midrule
\rowcolor{gray!25}&\textbf{FourGCD (Ours)} &        Estimated&\bf{71.0}&\bf{74.4}&\bf{69.2}\\
\bottomrule
\end{tabular}
}}

  \label{tab:estsupp}

\end{table}

\noindent\textbf{Comparison across fine-tuning with different numbers of blocks}. 
We evaluated SelEx, SimGCD, and FourEx with 1, 2, or 3 fine-tuned blocks (\cref{tab:finetune}). Increasing blocks sometimes can also degrade performance (SimGCD), while FourEx consistently outperforms all baselines, showing that gains come from frequency-based supervision and its effectiveness increased with more blocks.

\begin{table}[h!]
\caption{Performance across fine-tuning with different numbers of blocks on the CUB dataset. FourEx consistently improves upon its baseline SelEx. 
}
\centering
\resizebox{\columnwidth}{!}{%
\setlength{\tabcolsep}{7pt}{\small
\begin{tabular}{laccaccacca}
\toprule
&\multicolumn{3}{c}{\textbf{One block}}&\multicolumn{3}{c}{\textbf{Two blocks}}&\multicolumn{3}{c}{\textbf{Three blocks}}&\multicolumn{1}{c}{\textbf{Avg}}\\ 
\cmidrule(lr){2-4} \cmidrule(lr){5-7} \cmidrule(lr){8-10} \cmidrule(lr){11-11} 
\textbf{Method}  &\textbf{All}&\textbf{Known}&\textbf{Novel}&\textbf{All}&\textbf{Known}&\textbf{Novel}&\textbf{All}&\textbf{Known}&\textbf{Novel}&\textbf{All}\\
\midrule
SimGCD &60.3&65.6&57.7&	58.0 &57.6&58.1&56.6& 55.0 &57.5&58.3\\
SelEx &63.2 &69.3 &60.1&73.6 &75.3 &72.8& 76.4 &72.4 &78.4&71.1\\

\hline
\rowcolor{gray!25}FourEx & 71.4& 72.0 &71.1&78.3 &75.7 &79.6&80.2 &75.1 &82.7&76.6 \\

\bottomrule
\end{tabular}
}
}
\label{tab:finetune}
\end{table}

\section{Related Works}

\subsection{Novel Category Discovery.}
~proposed by Han \emph{et al.}~\cite{han2019learning} focuses on transferring classification knowledge from known to novel categories. Initial approaches \cite{han2019deep, hsu2017learning, hsu2019multi} adopted a two-stage methodology, where representation learning was confined to labeled data in the first stage, followed by the transfer of learned category structures to unknown categories in the subsequent stage. Recent advancements have shifted towards a more integrated, single-stage approach, \emph{eg},~\cite{fini2021unified, han2020automatically, zhao2021novel, zhong2021openmix, roy2022class, rizve2022towards}, where both labeled and unlabeled data are utilized concurrently for representation learning. Despite these developments, a fundamental challenge persists in the novel category discovery: the assumption of mutual exclusivity between known and novel categories. In our work, we focus on the pragmatic approach of uncovering both novel and known categories concurrently. 

\subsection{Generalized Category Discovery.}
~The task of Generalized Category Discovery (GCD) was formalized by Vaze~\etal~\cite{vaze2022generalized} and Cao~\etal~\cite{cao2021open}. GCD lies at the intersection of unsupervised and supervised learning. It leverages a small amount of labeled data along with a larger set of unlabeled data, where the unlabeled data can contain both known and novel categories. This makes GCD as a special case of self-supervised learning \cite{ouali2020overview, yang2022survey, rebuffi2020semi, oliver2018realistic, chapelle2009semi}. There are two prominent approaches for GCD: \textit{Prototype-based Methods:} These approaches leverage a set of pre-defined prototypes as reference points to guide category discovery in the unlabeled data. This can be achieved through techniques like nearest neighbor search or learning a distance metric that effectively separates known and unknown categories, \cite{hao2024cipr, chiaroni2023parametric, wen2022simple, an2023generalized}. 
\textit{Similarity-driven Clustering:} This approach exploits local similarities within the unlabeled data to form initial category clusters. This can be done through techniques like k-nearest neighbors or using mean-teacher frameworks to mitigate the issue of noisy pseudo-labels generated from similar information.
\cite{pu2023dynamic,zhang2022promptcal, hao2024cipr, chiaroni2023parametric, rastegar2023learn, banerjee2024amend, otholt2024guided} or by utilizing mean-teacher networks to address the challenges posed by noisy pseudo-labels~\cite{vaze2023clevr4,zhang2022promptcal, wen2022simple}. Nonetheless, Contrastive learning methods often struggle with fine-grained GCD due to aggressive augmentations overshadowing subtle category differences \cite{cole2022does}. 
Our work addresses this by leveraging the informative nature of high frequencies for fine-grained discrimination, while utilizing the lower frequencies for category discovery, enabling better handling of nuanced visual details. Several recent works explore alternative approaches for GCD. Hierarchical approaches proposed by Otholt~\etal~\cite{otholt2024guided} and Banerjee~\etal~\cite{banerjee2024amend} leverage neighborhood structures for refined category delineation. Rastegar~\etal~\cite{rastegar2023learn} introduced implicit category trees that facilitate hierarchical self-coding, maintaining category similarity across different levels. Choi~\etal~\cite{choi2024contrastive} focuses on robustness to noise by employing the mean shift algorithm for category discovery. Additionally, SPTNet \cite{wang2024sptnet} proposes a spatial prompt tuning method that incorporates spatial information from image data to focus better on specific object parts. Unlike these works, our method leverages weak supervision by focusing on high frequencies to distinguish samples from each other while focusing on low frequencies to categorize them.

\subsection{Fourier Transform Based Image Augmentation.}
~Fourier image augmentation has gained attention in computer vision due to its potential to enhance the robustness and generalizability of deep learning models. It leverages the Fourier transform to modify image data in the frequency domain, providing unique advantages over traditional spatial domain augmentations. Yang~\etal~\cite{yang2020fda} proposed Fourier Domain Adaptation (FDA), which applies the Fourier transform to both source and target domain images to align their frequency distributions. This improves classification accuracy significantly in domain adaptation scenarios by focusing on the low-frequency components that capture essential structural information while ignoring high-frequency noise. Sun~\etal~\cite{sun2022spectral} introduced FourierMix, a data augmentation technique that blends images in the frequency domain. This method creates new training examples by mixing different images' amplitude and phase spectra, leading to improved generalization performance on various benchmark datasets. Xu~\etal~\cite{xu2021fourier} used Fourier magnitude swaps between different samples to improve domain generalization. Distinct from these approaches our method uses the Fourier frequency for discovering novel categories in the context of generalized category discovery. Additionally, in addition to using the Fourier transform for data augmentation, we use it to specialize different parts of the latent dimension for different frequencies.
 
\section{Broader Impacts}
\textit{Reduced Data Labeling Costs:} GCD addresses the challenge of unlabeled data, which is the vast majority in the real world. By effectively discovering categories from this data, GCD has the potential to significantly reduce the computational and human effort required for manual data labeling. This can lead to faster development cycles and more efficient use of resources in various applications. \textit{Privacy-Preserving Applications:} Our specific approach utilizing the Fourier transform holds promise for applications requiring data privacy. Since the Fourier transform can blur data while preserving key features, it can be used in conjunction with GCD to discover categories from sensitive data without compromising its confidentiality. This could be beneficial in domains like healthcare or surveillance where privacy regulations are crucial.\\
\textit{Accelerating Pretraining of Large-Scale Models:}
Our motivation for introducing Fourier self-supervision is rooted in the $F$-principle, which suggests that neural networks tend to learn lower-frequency components of data before higher-frequency ones. Without explicit guidance, models may require extensive training to organically traverse this path. By directly aligning the learning dynamics with this natural frequency progression, our approach provides a principled shortcut. As a result, it has the potential to significantly reduce the computational cost and time associated with pretraining large models, contributing to more sustainable and efficient model development across the field.


\end{document}